\documentclass[12pt]{article}

\usepackage[margin=1in]{geometry}  
\usepackage{epsfig}
\usepackage{graphics,graphicx,color}
\usepackage{amssymb,amsfonts,amsthm,amsmath}
\usepackage{multirow, booktabs}
\usepackage{mathrsfs}
\usepackage{natbib}
\usepackage{slashbox}
\usepackage{longtable,booktabs}
\usepackage{arydshln}
\usepackage{paralist}

\newcommand{\bm}[1]{\mbox{\boldmath$ #1 $\unboldmath}}

\newcommand{\Exp}{\mathbb{E}}

\begin{document}


\title{Investigating the Robustness of Artificial Intelligent Algorithms with Mixture Experiments}

\author{Jiayi Lian, Laura Freeman, Yili Hong, and Xinwei Deng\footnote{Address for correspondence: Xinwei Deng, Associate Professor of Statistics, Co-Director of Statistics and Artificial Intelligence Laboratory at Virginia Tech (Email: xdeng@vt.edu).}\\
Department of Statistics, Virginia Tech, Blacksburg, VA 24061
}
\date{}
\maketitle

\begin{abstract}
Artificial intelligent (AI) algorithms, such as deep learning and XGboost, are used in numerous applications including computer vision, autonomous driving, and medical diagnostics.
The robustness of these AI algorithms is of great interest as inaccurate prediction could result in safety concerns and limit the adoption of AI systems.
In this paper, we propose a framework based on design of experiments to systematically investigate the robustness of AI classification algorithms.
A robust classification algorithm is expected to have high accuracy and low variability under different application scenarios.
The robustness can be affected by a wide range of factors such as the imbalance of class labels in the training dataset, the chosen prediction algorithm, the chosen dataset of the application, and a change of distribution in the training and test datasets.
To investigate the robustness of AI classification algorithms, we conduct a comprehensive set of mixture experiments to collect prediction performance results.
Then statistical analyses are conducted to understand how various factors affect the robustness of AI classification algorithms.
We summarize our findings and provide suggestions to practitioners in AI applications.

\end{abstract}


\newpage
\section{Introduction}
\subsection{Background}
Machine learning (ML) and deep learning (DL) algorithms are widely used in many artificial intelligent (AI) applications such as computer vision \citep{hashimoto2019computer}, autonomous driving \citep{tian2018deeptest}, and medical diagnostics \citep{quer2017augmenting}.
There is a growing interest in investigating the robustness of AI algorithms \citep{dietterich2017steps, silva2020opportunities, hamon2020robustness}, as it is highly related to the safety of AI systems~\citep{Amodeietal2016}.
The robustness of AI algorithms often concerns how the performance of the algorithm may change when the training and test datasets are of certain differences \citep{xu2012sparse}.
For example, adding noise into the training data can change the prediction performance of the AI algorithms to some extent.
The robustness of the AI algorithms heavily relies on how the algorithm is trained and how the algorithm is evaluated on the test dataset of interest.
In this work, we focus on investigating the robustness of classification algorithms as many artificial intelligent (AI) systems involve classification.
The proposed framework, however, can be extended to study the robustness for AI algorithms beyond classification.

A robust classification algorithm is expected to have high accuracy and low variability under various scenarios.
The robustness of the classification algorithm can be affected by a wide range of factors, such as the composition of training dataset, the change of distribution in the training and test datasets, the chosen prediction algorithm, and so on.
In the training stage, AI algorithms often rely on a large number of data points.
However, in many real-world applications, data for classification can be highly imbalanced in which data points from a particular class label overwhelmingly dominate data points from other class labels.
It is crucial to understand how the performance of AI classification algorithms is affected by the class label imbalance in the training data.
Moreover, when the distribution of test data deviates away from the distribution of the training data,
such distribution changes can also largely affect the robustness of the classification algorithms \citep{kuleshov2015calibrated}. This lack of robustness has direct application to deployed algorithms used on data sets whose composition may differ from the original training data set.


In this paper, we use the idea of design of experiments \citep{wu2011experiments} to systematically investigate the robustness of classification algorithms
with respect to: (1) the class label imbalance in the training data; and (2) the distribution change between training and test data regarding the proportions of class labels.
In particular, our key idea is to use the mixture experimental design \citep{cornell2011experiments} for the proportions of class labels in the training data,
such that the robustness of the AI classification algorithms can be investigated in a systematic manner.
Here the robustness of a classification algorithm is characterized by the mean and standard deviation of the areas under the receiver operating characteristic curves (AUC) for each class label \citep{YuanBar-Joseph2019}.
We also study a wide range of other factors that can affect the robustness, including the distributional changes between the training data and the test data, different classification algorithms, and datasets from different applications.
The convolutional neural network (CNN) \citep{kim2014convolutional} and XGBoost \citep{chen2015xgboost} are adopted as two representative AI classification algorithms.
Based on the classification performance results collected from the designed mixture experiments,
statistical analyses are conducted to reveal some interesting findings on the robustness of AI classification algorithms.
It is worth to remarking that the proposed framework can be extended to including more factors for investigating the robustness of the AI algorithms,
and can also be used to study other characteristics of the AI algorithms, such as the transparency of the AI systems \citep{hamon2020robustness}.

\subsection{Related Literature}
In a typical mixture experiment \citep{cornell2011experiments}, the design variables under study are the proportions of mixture components in a blend, with their summation equal to unity.
A common objective is to investigate how the changes in the proportions of mixture components would affect the response outcomes \citep{PiepelCornell1994}.
The mixture experiments have been used in many agriculture and engineering applications \citep{kang2011design, kang2016comparing, shen2020additive}.
In our work, we adopt the idea of mixture experiments to study how the proportions of class labels would affect the performance of the AI classification algorithms.
Among various AI classification algorithms, we consider the convolutional neural network (CNN) and the XGboost as the two representative algorithms.
The CNN is a popular deep learning algorithm for classification, where the input variables are tensors \citep{goodfellow2016deep}.
Generally, the CNN uses convolution in place of general matrix multiplication in its perception layers.
In contrast, XGboost is one efficient classification algorithm when the input contains numerical and categorical variables.
The XGBoost stands for the Extreme Gradient Boosting, which is a parallel tree boosting method with efficient gradient-based optimization.
Both CNN and XGboost are widely used in various application areas \citep{parsa2020toward}.
With the consideration of mixture experiments for the proportions of class labels,
our approach is to provide a comprehensive understanding on the performance of these two classification algorithms.

The robustness of AI algorithms has attracted a large amount of attention in the machine learning community \citep{tsipras2018robustness}.
There is a broad spectrum regarding the robustness of AI algorithms, including
adversarial robustness, robust learning, robust models, robustness to distributional shift, etc.
In the robust learning, a key interest is to redesign the learning procedure so that the algorithm is robust against malicious actions \citep{madry2017towards, zantedeschi2017efficient}.
Robust procedures include better training procedures against adversarial examples, as well as a better mathematical foundation of the algorithms by adopting techniques from statistics and optimization, such as robust statistic inference \citep{huber2004robust, lozano2013robust} and robust optimization \citep{ben2009robust}.
In the area of deep learning, some research work has focused on investigating the adversarial robustness \citep{dvijotham2018dual, xiang2018output, gehr2018ai2, wong2018provable}.
It is known that neural network models are vulnerable to adversarial examples.
That is, perturbing inputs that are very similar to some regular inputs could result in the output being dramatically different \citep{szegedy2013intriguing}.
For example, \cite{tjeng2017evaluating} proposed a mixed integer programming method to examine the vulnerability of neural networks to such adversarial examples.
A useful framework for certifying the robustness of CNNs against adversarial examples is discussed in \cite{boopathy2019cnn}.
The scope of our work is to use a framework of experimental design to investigate the robustness of AI algorithms
against the class label imbalance in the training data and the distribution change on proportions of class labels between the training and test data.

\subsection{Overview}
The rest of the paper is organized as follows.
Section~\ref{sec:prop.framework} describes the proposed framework, including the response variables, the experimental factors, design runs, data collection and the modeling method. Section~\ref{sec:model.analysis} reports the analysis results of experimental data.
Section~\ref{sec:discussion} contains some concluding remarks.

\section{The Proposed Framework}\label{sec:prop.framework}
\subsection{Design Factors and Response Variables}
Let us consider a multi-class classification with $m$ classes (labels).
The accuracy of a classification algorithm can be measured by various performance measures, such as the misclassification error, false positive rate (FPR), $F_{1}$ score, etc.
Note that the classification rule of a classification algorithm often depends on the setting of the classification threshold \citep{YuanBar-Joseph2019}.
Among various performance measures, the area under the receiver operating characteristic curve (AUC) provides an aggregate measure of prediction accuracy for all possible thresholds.
For a multi-class classification problem, one can obtain the AUC for each class label.
To characterize the robustness of a classification algorithm, we consider two types of performance measures based on the AUC's.
The first one is the sample mean of the AUC values of $m$ classes, and the second one is the logarithm of the standard deviation of the AUC values from $m$ classes.


To investigate the robustness of the AI classification algorithms,
we consider a mixture experiment on proportions of class label in the training dataset
with several covariate variables such as different algorithms and different datasets of interest.
Specifically, let $(x_{1}, x_{2}, \ldots, x_{m})$ be the proportions of class labels $\{1, \ldots, m\}$ in the training dataset.
Note that $x_{1} + x_{2}+ \cdots + x_{m} = 1$ and $x_{j} \in [0,1]$ for $j = 1,\ldots, m$.
Suppose that there are $h$ covariate variables, $z_{1}, \ldots, z_{h}$, for the mixture experiment.
In our specific experimental setting, we consider two covariate variables $z_{1}$ and $z_{2}$, each with two levels.
In particular, $z_{1}$ is a two-level factor that represents two different algorithms: the CNN algorithm and the XGboost algorithm.
That is,
\begin{align*}
z_{1}=
\begin{cases}
    1, & \text{if the XGboost algorithm is used},\\
    0, & \text{if the CNN algorithm is used}.
\end{cases}
\end{align*}
The $z_{2}$ variable is a two-level factor that represents two different dataset of interests: the KEGG dataset and the Bone Marrow dataset.
That is,
$$
z_{2}=
    \begin{cases}
    1, & \text{if the KEGG dataset is used},\\
    0, & \text{if the Bone Marrow dataset is used}.
    \end{cases}
$$
The above two datasets are used in \cite{YuanBar-Joseph2019} for the classification of the single cell RNA sequencing (scRNA-seq) expression.
In \cite{YuanBar-Joseph2019}, the scRNA-seq data are converted into the normalized empirical probability distribution functions (NEPDF) between gene pairs as the input matrix
and the CNN is applied for classification.
The KEGG dataset is derived from the Kyoto Encyclopedia of Genes and Genomes (KEGG) as pathway datasets of study \citep{wixon2000kyoto}.
The Bone Marrow dataset denotes the bone marrow derived  macrophage scRNA-seq as the dataset of study \citep{orlic2001bone}.
The KEGG dataset contains $92{,}472$ observations and the Bone Marrow dataset has $80{,}253$ observations.
Both datasets have $m=3$ classes with class proportions equally distributed, i.e., each class contains one third of observations.
The details about the two datasets and their pre-processing can be found in  \citep{YuanBar-Joseph2019}.

\subsection{Design Construction and Runs}
\begin{figure}
    \centering
    \includegraphics[width=0.6\textwidth]{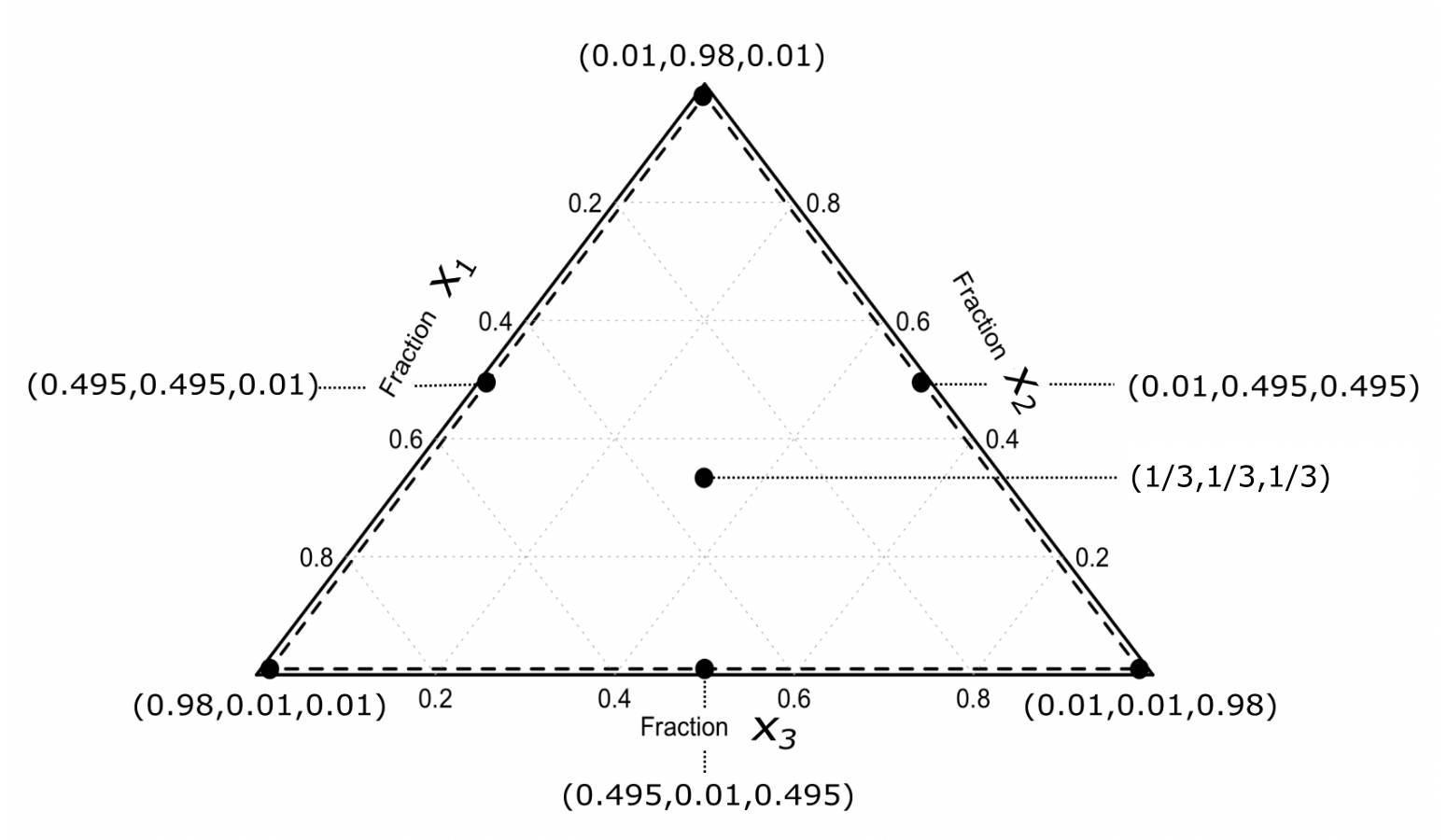}
    \caption{Illustration of seven settings of proportions of class labels in the training dataset.
    The dashed lines illustrate the restriction on the minimal value of a class proportion larger than 0.01, i.e., $x_{j} \ge 0.01$.}
    \label{fig:mixture-design}
\end{figure}

Because the datasets of interest have three class labels, we consider a modified simple centroid design for mixture experiment for $x_{1}, x_{2}, \ldots, x_{m}$ with $m = 3$.
A simple centroid design usually has $2^{m} - 1$ points, including $m$ pure components (e.g., (0,0,1)), $\tbinom{m}{2}$ binary points (e.g., (0,1/2,1/2)) and $\tbinom{m}{3}$ ternary mixture (e.g., (1/3,1/3,1/3)).
When $m = 3$, there are seven different settings of proportions of class labels for the training dataset,
as shown in Figure~\ref{fig:mixture-design}.
With the consideration of the covariate variables $z_{1}$ and $z_{2}$ with two-levels,
we use the cross-array between the mixture design of $x_{1}, x_{2}, x_{3}$ and the full factorial design of $z_{1}$ and $z_{2}$.
The design matrix of our proposed method is shown in Table~\ref{tab:design-table}.
\begin{table}
\begin{center}
\caption{The 28-run mixture design with 2 covariate factors.}\label{tab:design-table}
\vspace{1.5ex}
\begin{tabular}{lrrrrr|clrrrrr}\hline\hline
Run & $x_{1}$ & $x_{2}$ & $x_{3}$ & $z_{1}$ & $z_{2}$ &&
Run& $x_{1}$ & $x_{2}$ & $x_{3}$ & $z_{1}$ & $z_{2}$ \\\hline
1 &  0.01 &  0.01 &  0.98 & 1 & 1 & & 15 & 0.01 &  0.01 &  0.98 & 0 & 1 \\
2 &  0.01 &  0.98 &  0.01 & 1 & 1 & & 16 & 0.01 &  0.98 &  0.01 & 0 & 1 \\
3 &  0.98 &  0.01 &  0.01 & 1 & 1 & & 17 & 0.98 &  0.01 &  0.01 & 0 & 1 \\
4 &  0.01 & 0.495 & 0.495 & 1 & 1 & & 18 & 0.01 & 0.495 & 0.495 & 0 & 1 \\
5 & 0.495 &  0.01 & 0.495 & 1 & 1 & & 19 &0.495 &  0.01 & 0.495 & 0 & 1 \\
6 & 0.495 & 0.495 &  0.01 & 1 & 1 & & 20 &0.495 & 0.495 &  0.01 & 0 & 1 \\
7 &  1/3 &  1/3 &  1/3 & 1 & 1 & & 21 & 1/3 &  1/3 &  1/3 & 0 & 1 \\
8 &  0.01 &  0.01 &  0.98 & 1 & 0 & & 22 & 0.01 &  0.01 &  0.98 & 0 & 0 \\
9 &  0.01 &  0.98 &  0.01 & 1 & 0 & & 23 & 0.01 &  0.98 &  0.01 & 0 & 0 \\
10 & 0.98 &  0.01 &  0.01 & 1 & 0 & & 24 & 0.01 &  0.01 &  0.98 & 0 & 0 \\
11 & 0.01 & 0.495 & 0.495 & 1 & 0 & & 25 & 0.01 & 0.495 & 0.495 & 0 & 0 \\
12 &0.495 &  0.01 & 0.495 & 1 & 0 & & 26 &0.495 &  0.01 & 0.495 & 0 & 0 \\
13 &0.495 & 0.495 &  0.01 & 1 & 0 & & 27 &0.495 & 0.495 & 0.01  & 0 & 0 \\
14 & 1/3 &  1/3 &  1/3 & 1 & 0 & & 28 & 1/3 &  1/3 & 1/3  & 0 & 0 \\ \hline\hline
\end{tabular}
\end{center}
\end{table}

Note that our goal is to investigate the robustness of AI classification algorithms on their prediction performance.
For the experimental setting of $x_{1}, x_{2}, x_{3}$, the proportions of class labels in the training dataset,
it is not practical to set the proportion of a class label to be zero, i.e., $x_{j} = 0$ for some $j$.
Under this consideration, we modify the simple centroid design to restrict the minimum value of class proportion to be larger than 0.01, i.e., $x_{j} \ge 0.01$, which is illustrated by the dashed lines in Figure~\ref{fig:mixture-design}.


To calculate the classification accuracy, a test dataset is often needed.
The proportions of class labels in the test dataset also can have an effect on the classification performance of the algorithms.
Often time, we assume that the training and test datasets have the same distributions on the proportion of class labels.
In practice, the distribution of proportions of class labels in the test dataset can be different from that in the training dataset.
To comprehensively evaluate the robustness of the AI classification algorithm,
we will alter the distribution of proportions of class labels in the test dataset to be possibly different from the distribution of class proportions in the training dataset.
Specifically, three scenarios for the test dataset are considered, which are listed as follows.
\begin{itemize}
    \item {\bf Balanced Scenario}: The proportions of class labels in the test dataset are equal in each class, regardless of the proportions of class labels in the training dataset.

    \item {\bf Consistent Scenario}: The proportions of class labels in the test dataset are the same as the proportions of class labels in the training dataset.

    \item {\bf Reverse Scenario}: The proportion of class labels in the test dataset is in a reverse pattern as the proportion of class labels in the training dataset.
\end{itemize}
In the Reverse Scenario, if the proportion of a class label is high in the training dataset, then the proportion of this class label is set to be low in the test dataset.
Table~\ref{tab:test-scenario} shows the three scenarios of proportions of class labels between the training and test dataset.

\begin{table}
\begin{center}
\caption{Three scenarios of proportions of class labels for forming the test dataset.}\label{tab:test-scenario}
\vspace{1.5ex}
\begin{tabular}{lrrr clrrr}
\hline\hline
\multicolumn{8}{c}{Balanced Scenario}\\
Training & $x_{1}$ & $x_{2}$ & $x_{3}$  && Test & $x_{1}$ & $x_{2}$ & $x_{3}$  \\\hline
&  0.01 &  0.01 &  0.98  & &   &  1/3 &  1/3 &  1/3  \\
&  0.01 &  0.98 &  0.01  & &   &  1/3 &  1/3 &  1/3 \\
&  0.98 &  0.01 &  0.01  & &   &  1/3 &  1/3 &  1/3 \\
&  0.01 & 0.495 & 0.495  & &   &  1/3 &  1/3 &  1/3 \\
& 0.495 &  0.01 & 0.495  & &   &  1/3 &  1/3 &  1/3  \\
& 0.495 & 0.495 &  0.01  & &   &  1/3 &  1/3 &  1/3 \\
&  1/3 &  1/3 &  1/3 & &   &  1/3 &  1/3 &  1/3 \\
\hline
\multicolumn{8}{c}{Consistent Scenario}\\
Training & $x_{1}$ & $x_{2}$ & $x_{3}$  && Test & $x_{1}$ & $x_{2}$ & $x_{3}$  \\\hline
&  0.01 &  0.01 &  0.98  & &   & 0.01 &  0.01 &  0.98  \\
&  0.01 &  0.98 &  0.01  & &   & 0.01 &  0.98 &  0.01 \\
&  0.98 &  0.01 &  0.01  & &   & 0.98 &  0.01 &  0.01 \\
&  0.01 & 0.495 & 0.495  & &   & 0.01 & 0.495 & 0.495 \\
& 0.495 &  0.01 & 0.495  & &   &0.495 &  0.01 & 0.495  \\
& 0.495 & 0.495 &  0.01  & &   &0.495 & 0.495 &  0.01 \\
&  1/3 &  1/3 &  1/3  & &   &  1/3 &  1/3 &  1/3 \\
\hline
\multicolumn{8}{c}{Reverse Scenario}\\
Training & $x_{1}$ & $x_{2}$ & $x_{3}$  && Test & $x_{1}$ & $x_{2}$ & $x_{3}$  \\\hline
&  0.01 &  0.01 &  0.98  & &   & 0.495 &  0.495 &  0.01  \\
&  0.01 &  0.98 &  0.01  & &   & 0.495 &  0.01 &  0.495 \\
&  0.98 &  0.01 &  0.01  & &   & 0.01 &  0.495 &  0.495 \\
&  0.01 & 0.495 & 0.495  & &   & 0.98 & 0.01 & 0.01 \\
& 0.495 &  0.01 & 0.495  & &   &0.01 &  0.98 & 0.01  \\
& 0.495 & 0.495 &  0.01  & &   &0.01 & 0.01 &  0.98 \\
\hdashline
&         &     &        & &   &  $^\ast$0.01 &  0.01 &  0.98 \\
&  1/3   & 1/3 &  1/3  & &   &  $^\ast$0.01 &  0.98 &  0.01\\
&         &     &        & &   &  $^\ast$0.98 &  0.01 &  0.01 \\
 \hline\hline
\multicolumn{8}{l}{$^\ast$Randomly select one from the three proportion settings.}\\
\end{tabular}
\end{center}
\end{table}

Note that for all experimental runs,
we keep the size of the training dataset to the same with $n_{tr} = 10\% \mbox{\textit{ of the full dataset}}$,
and keep the size of the test dataset to be the same with $n_{ts} = 25\% \mbox{\textit{ of the full dataset}}$. Here, we make the size of the test dataset larger than that of the training dataset for the following considerations.
First, it allows the evaluation of classification performance at the test datasets being more valid, avoid the potential variation due to the small sample size.
Second, a relatively smaller size of the training dataset could better serve the purpose of investigating the robustness of the algorithms.
In the full datasets of the KEGG (of the size $92{,}472$ observations) and Bone Marrow (of the size $80{,}253$ observations),
the class proportions are equally distributed, i.e., each label having one third of the full data.
In each experimenting run, when forming the training dataset, we randomly sample with replacement from the full dataset in each class label
based on the proportion setting $(x_{1}, x_{2}, x_{3})$ for training.
The use of sampling with replacement for forming the training dataset is to simulate the possible duplicated data points in the real practice \citep{li2020datafiltering}.
For forming the test dataset,
we randomly sample without replacement from the remaining data in each class label based on the proportion setting $(x_{1}, x_{2}, x_{3})$ for test. We use the sampling without replacement to make each data point unique in the test dataset.

To illustrate how data points with its percentage for each label are composed in both training and test datasets, we show a toy example as follows.
Suppose that there are $N = 10000$ observations in the full dataset with three class proportions equally distributed.
Then we set the size of training dataset to be fixed at $n_{tr} = 10\%\cdot N = 1000$,
while the size of test dataset is set to be $n_{ts} = 25\%\cdot N = 2500$.
Assume that in one experimental run, the proportions of class labels in the training dataset is designed to be $(x_{1}, x_{2}, x_{3}) = (0.01,0.01,0.98)$,
and the proportions of class labels in the test dataset is designed to be $(x_{1}, x_{2}, x_{3}) = (0.01,0.01,0.98)$.
Based on the full dataset, the training dataset is composed by randomly sampling $(10, 10, 980)$ observations from the three classes of data points, respectively.
Then using the remaining data points, the test dataset is composed by randomly sampling $(25,25,2450)$ observations from the three classes of the remaining data points, respectively.


\subsection{Data Collection and Modeling Method}
As shown in Table~\ref{tab:design-table}, there are 28 treatments (i.e., runs) in the proposed design under each scenario of test data.
For each treatment, we conduct three replications.
All experiments were carried out on a DGX-2 machine which is a product of NVIDIA focusing on deep learning implementation.
On average, each experiment takes around 30 minutes in computation.
The outcome of an experiment is the AUC value of each class on the test dataset
by a selected classification algorithm (CNN or XGboost), which is estimated based on the training data.
Note that for the classification algorithms, the input of the CNN model is the matrix of the normalized empirical probability distribution functions (NEPDF) between gene pairs,
which is an image-type input. We adopt the same implementation of the CNN as in \cite{YuanBar-Joseph2019}.
While in the implementation of XGboost, the inputs contain the column sum, row sum, and the trace of the matrix of NEPDFs.
In fact, column sums and row sums are the marginal empirical distribution of each gene in the pair, and the trace is also a summary of the matrix.

Based on the AUC value of each class, denoted as $\eta_{1}, \ldots, \eta_{m}$, then the averaged AUC (mean AUC) can be obtained as
$$\bar{\eta} = \frac{1}{m}\sum_{j=1}^{m} \eta_{j} \ ,$$
and the logarithm of the standard deviation (Log SD) can be obtained as
$$\textrm{Log SD} = \log\left(\left[ \frac{1}{m-1} \sum_{j=1}^{m} (\eta_{j} - \bar{\eta})^{2}\right]^{1/2}\right) \,.$$


When the outcomes of designed experiments (i.e., the collected data) are obtained,
our key interest is to conduct proper statistical modeling to understand how the design factors (class label proportions $x_{1}, \ldots, x_{m}$, the algorithms, and the datasets of interest) affect the response (the classification accuracy).
Note that we have three test scenarios, the Consistent Scenario, the Balanced Scenario, and the Reverse Scenario.
We will conduct the statistical analysis of experimental data separately for each scenario.

Given a given test scenario,
we denote $(x_{i1}, x_{i2}, \ldots, x_{im})$ as the proportions of class labels of the training dataset for the $i$th run,
and $z_{ik}$ to be the level of covariate variable $z_{k}$ for the $i$th run.
Let $y_i$ be the corresponding response variable, which can be the mean AUC or the Log SD.
Let $n$ be the sample size of the collected data under a given test scenario.
To analyze the collected data $(y_i, x_{i1}, x_{i2},\ldots, x_{im}, z_{i1}, \ldots, z_{ih})$, we consider the following regression model,
\begin{align}\label{eqn:reg.model}
y & = \sum_{j=1}^{m} ( \beta_j + \sum_{k=1}^{h} \gamma_{kj} z_{k} ) x_{j} +  \sum_{j < j'} \beta_{j j'}x_{j}x_{j'} + \sum_{k < k'}\delta_{kk'} z_{k}z_{k'} +\epsilon \nonumber \\
&= \sum_{j=1}^{m} \beta_j x_{j} +  \sum_{j < j'} \beta_{j j'}x_{j}x_{j'} + \sum_{k=1}^{h} \sum_{j=1}^{m} \gamma_{kj} z_{k} x_{j} + \sum_{k < k'}\delta_{kk'} z_{k}z_{k'} +\epsilon,
\end{align}
where $\beta_{j}$'s and $\beta_{j j'}$'s are regression coefficients for the main and interaction terms of the label proportions, respectively.
The $\gamma_{kj}$'s are the regression coefficients for the interaction between class label proportions and the covariate variables,
and $\delta_{kk'}$'s are the regression coefficients for the interactions of the covariate variables.
The $\epsilon \sim \textrm{N}(0, \sigma^{2})$ is the error term.
Here, the model does not include the high-order interaction terms to keep the model parsimonious.

Denote $\bm \beta$ to be the vector of coefficients including $\beta_{0}$, $\beta_{j}$'s and $\beta_{j j'}$'s.
Similarly, we define $\bm \gamma$ and $\bm \delta$ to be the vectors of coefficients for $\gamma_{kj}$'s and $\delta_{kk'}$'s, respectively.
For the parameter estimation, we adopt the maximum likelihood estimation, which is equivalent to the least squares estimation in our problem.
That is,
\begin{align*}
\footnotesize (\hat{\bm \beta}, \hat{\bm \gamma}, \hat{\bm \delta} ) = \arg \min_{\bm \beta, \bm \gamma, \bm \delta}
\sum_{i=1}^{n} \left (y_{i} - \sum_{j=1}^{m} \beta_j x_{ij} - \sum_{j < j'} \beta_{j j'}x_{ij}x_{ij'} - \sum_{k=1}^{h} \sum_{j=1}^{m} \gamma_{kj} z_{k} x_{ij} - \sum_{k < k'}\delta_{kk'} z_{ik}z_{ik'} \right )^{2}.
\end{align*}

Note that the model in \eqref{eqn:reg.model} does not include an intercept term because the class label proportions sum up to one, i.e., $x_{1}+\cdots+x_{m} = 1$.
The model in \eqref{eqn:reg.model} also does not include the quadratic term of label proportions
since it can be linearly expressed by its main effect and its interactions with other class label proportions, i.e.,
$x_{j}^2 = x_{j} - \sum_{j'\ne j } x_{j} x_{j'}$.
Moreover, the model in \eqref{eqn:reg.model}  does not include the main effect of the covariate factor since it can be linearly expressed by the interactions between class label proportions and the covariate variables, i.e., $z_{k} = z_{k}x_{1} + \cdots + z_{k}x_{m}$.
To make proper inference on the contribution of $z_{k}$ in the estimated model,
one could include $z_{k}$ into the regression model in \eqref{eqn:reg.model}, which leads to the following terms in the model,
\begin{align}\label{eqnz-infer}
& \gamma_{k1} z_{k} x_{1} + \gamma_{k2} z_{k} x_{2} + \cdots + \gamma_{km} z_{k} x_{m} + \gamma_{k} z_{k} \nonumber \\
& = (\gamma_{k1} + \gamma_{k}) z_{k} x_{1} + (\gamma_{k2} + \gamma_{k}) z_{k} x_{2} + \cdots + (\gamma_{km} + \gamma_{k}) z_{k} x_{m}.
\end{align}
By imposing the sum-to-zero constraint \citep{wu2011experiments} for model identifiability, we have
\begin{align*}
(\gamma_{k1} + \gamma_{k}) + \cdots + (\gamma_{km} + \gamma_{k}) = 0
\Rightarrow \gamma_{k} = -\frac{1}{m} (\gamma_{k1} + \gamma_{k2} + \cdots + \gamma_{km} ).
\end{align*}
In this sense, we make inference on $z_{k}$ based on the linear combination of $\hat{\gamma}_{kj}$'s through $\frac{1}{m} \sum_{j=1}^{m} \hat{\gamma}_{kj}$.
It is worth to pointing out that one could also impose the baseline constraint, such as $\gamma_{k} = 0$ in \eqref{eqnz-infer},
then it becomes the original model in \eqref{eqn:reg.model}.


Moreover, based on the estimated model,
it is useful to quantify the impact of the predictor variables to the response.
Here we adopt the SHAP (SHapley Additive exPlanations) approach \citep{lundberg2017unified}  to quantitatively assess the impact of predictor variables to the response.
The Shapley value, based on the cooperative game theory \citep{shapley1953value}, can be applied in a wide variety of models and is not affected by the unit of measurement.
The SHAP method assigns each predictor variable a Shapley value of importance for the predictive model.
For the linear model in \eqref{eqn:reg.model}, the SHAP has an explicit form.
The detailed explanations of the Shapely formula under the linear model can be found in the appendix.
Denote $\phi_{i}^{(x_{j})}$ as the importance of the label proportion variable $x_{j}$ to the model output for individual observation $i$.
Similarly, we can define $\phi_{i}^{(z_{k}x_{j})}$, $\phi_{i}^{(x_{j}x_{j'})}$, $\phi_{i}^{(z_{k}z_{k'})}$.
For the $i$th observation in our proposed model,
the importance of predictor variables $x_{j}$'s, $x_{j}x_{j'}$'s, $z_{k}x_{j}$'s, and $z_{k}z_{k}'s$ have the following forms:
\begin{align*}
\phi_{i}^{(x_{j})} & =  \beta_{j} \Big( x_{ij} - \frac{1}{n}\sum_{i=1}^n x_{ij} \Big),  \quad j = 1, \ldots, m; \\
\phi_{i}^{(x_{j} x_{j'})} & =  \beta_{jj'} \Big( x_{ij}x_{ij'} - \frac{1}{n}\sum_{i=1}^n x_{ij}x_{ij'} \Big),  \quad j < j', j = 1, \ldots, m, j' = 1, \ldots, m;  \\
\phi_{i}^{(z_{k} x_{j})} & =  \gamma_{kj} \Big( z_{ik}x_{ij'} - \frac{1}{n}\sum_{i=1}^n z_{ik}x_{ij} \Big),  \quad k = 1, \ldots, h, j = 1, \ldots, m;  \\
\phi_{i}^{(z_{k} x_{j})} & =  \delta_{kk'} \Big( z_{ik}z_{ik'} - \frac{1}{n}\sum_{i=1}^n z_{ik}z_{ik} \Big),   \quad k < k', k = 1, \ldots, h, j' = 1, \ldots, h.
\end{align*}
The SHAP can also be used to quantify the overall impact of each predictor variable
by using the average absolute impact on model output magnitude as the evaluation metric; i.e.,
\begin{align}
\label{eqn:shap-1} \phi^{(x_{j})} & = \frac{1}{n} \sum_{i=1}^{n} | \phi_{i}^{(x_{j})} |,  \quad j = 1, \ldots, m; \\
\label{eqn:shap-2}  \phi^{(x_{j} x_{j'})} & = \frac{1}{n} \sum_{i=1}^{n} |\phi_{i}^{(x_{j} x_{j'})}|,  \quad j < j', j = 1, \ldots, m, j' = 1, \ldots, m;  \\
\label{eqn:shap-3}  \phi^{(z_{k} x_{j})} & = \frac{1}{n} \sum_{i=1}^{n} |\phi_{i}^{(z_{k} x_{j})}|,  \quad k = 1, \ldots, h, j = 1, \ldots, m;  \\
\label{eqn:shap-4}  \phi^{(z_{k} x_{j})} & = \frac{1}{n} \sum_{i=1}^{n} |\phi_{i}^{(z_{k} x_{j})}|,   \quad k < k', k = 1, \ldots, h, j' = 1, \ldots, h.
\end{align}
These metrics, $\phi^{(x_{j})}$'s, $\phi^{(x_{j} x_{j'})}$'s,$\phi^{(z_{k} x_{j})}$'s, and $\phi^{(z_{k} x_{j})}$'s will be used to assess and compare these impacts to the response in the next section.

\section{Data Analysis}\label{sec:model.analysis}
In this section, we conduct the modeling and data analysis when the response variable is the mean AUC and the Log SD, respectively.
In our analysis with $m=3$ class labels and $h=2$ covariate variables,
the collected data of experiments are $(y_{i}, x_{i1}, x_{i2}, x_{i3}, z_{1i}, z_{2i}), i = 1, \ldots, n$.
Then the model in \eqref{eqn:reg.model} is expressed as,
\begin{align}\label{eqn:reg.model-3}
y_i= & \beta_1 x_{i1} + \beta_2 x_{i2} +\beta_3 x_{i3} + \beta_{12}x_{i1}x_{i2} + \beta_{13}x_{i1}x_{i3} + \beta_{23}x_{i2}x_{i3} \\\nonumber
 & + \gamma_{11}z_{i1}x_{i1} + \gamma_{12}z_{i1}x_{i2} + \gamma_{13}z_{i1}x_{i3} +  \gamma_{21}z_{i2}x_{i1} + \gamma_{22}z_{i2}x_{i2} + \gamma_{23}z_{i2}x_{i3} + \delta_{12}z_{i1}z_{i2} +\epsilon_i.
\end{align}

\subsection{Data Visualization} \label{sec: data-visualization}

\begin{figure}
\begin{center}
\includegraphics[width=0.95\textwidth]{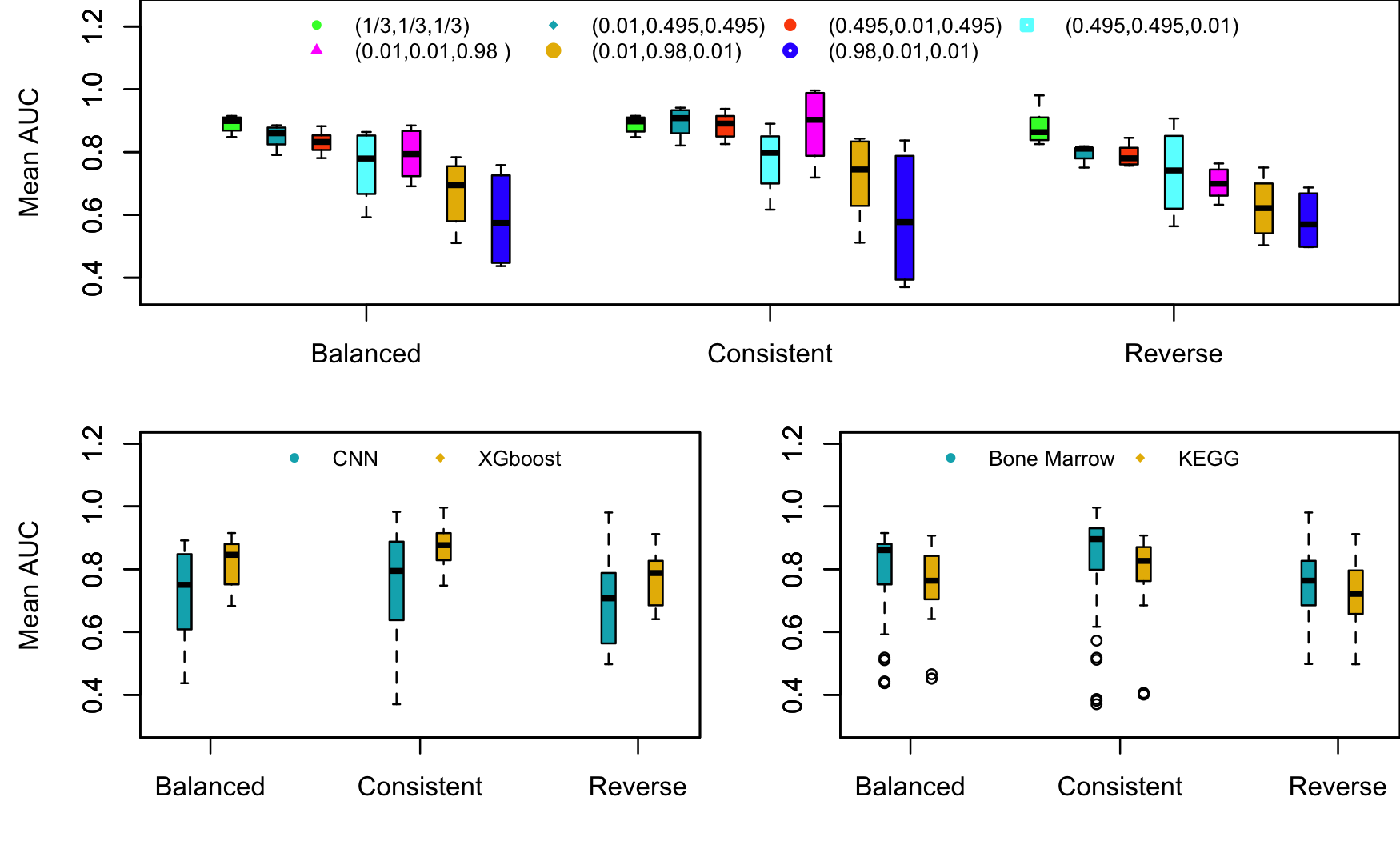}
\caption{Boxplots of the mean AUC response under different levels of covariate variable and different combinations of label proportions for the three test scenarios.}\label{fig:mean.auc}
\end{center}
\end{figure}

We first visualize the data from the experimental results for the mean AUC response and the Log SD response, respectively.
Figure~\ref{fig:mean.auc} displays the boxplots of the mean AUC response under different levels of covariate variables and different combinations of label proportions under the three test scenarios.
It is seen that for covariate variable $z_{1}$,  there is a clear difference between the CNN algorithm ($z_{1} = 0$) and the XGboost algorithm ($z_{1}=1$).
For the three test scenarios, it appears that the XGboost algorithm gives a higher value of mean AUC than the CNN algorithm.
Such a pattern is more evident under the Consistent Scenario in comparison with other scenarios.
But for covariate variable $z_{2}$, there is not a clear difference on the mean AUC between using the KEGG dataset ($z_{2}=0$) and using the Bone Marrow data ($z_{2} = 1$),
although using the KEGG dataset gives a slightly smaller Log SD than using the Bone Marrow dataset.
A possible explanation is that the two datasets are of similar data characteristics with respect to the overall classification accuracy,
as described in \citep{YuanBar-Joseph2019}.

For the boxplots of the mean AUC response under different combinations of label proportions,
it is seen that under the label proportions are balanced (i.e., $(x_{1}, x_{2}, x_{3}) = (1/3, 1/3, 1/3)$), the performance of mean AUC is usually better than those under other settings.
It implies that the balance of class label proportions in the training dataset plays an important role for AI classification algorithms to achieve robustly good accuracy.
When there are two classes dominating (e.g., $(x_{1}, x_{2}, x_{3}) = (0.01, 0.495, 0.495)$),
it is observed that the balance between class label 2 and label 3 ($x_2$ and $x_3$) gives better performance of mean AUC than the balance between class label 1 and label 2 ($x_1$ and $x_2$).
When there is only one class dominating (e.g., $(x_{1}, x_{2}, x_{3}) = (0.01, 0.01, 0.98)$),
we observe that the performance under the class label 3 dominating is better than the performance under the class label 1 or label 2 dominating.
This interesting pattern implies that the class label 3 plays a most substantial role among the three classes for the classification accuracy in terms of the mean AUC.

\begin{figure}
\begin{center}
\includegraphics[width=0.95\textwidth]{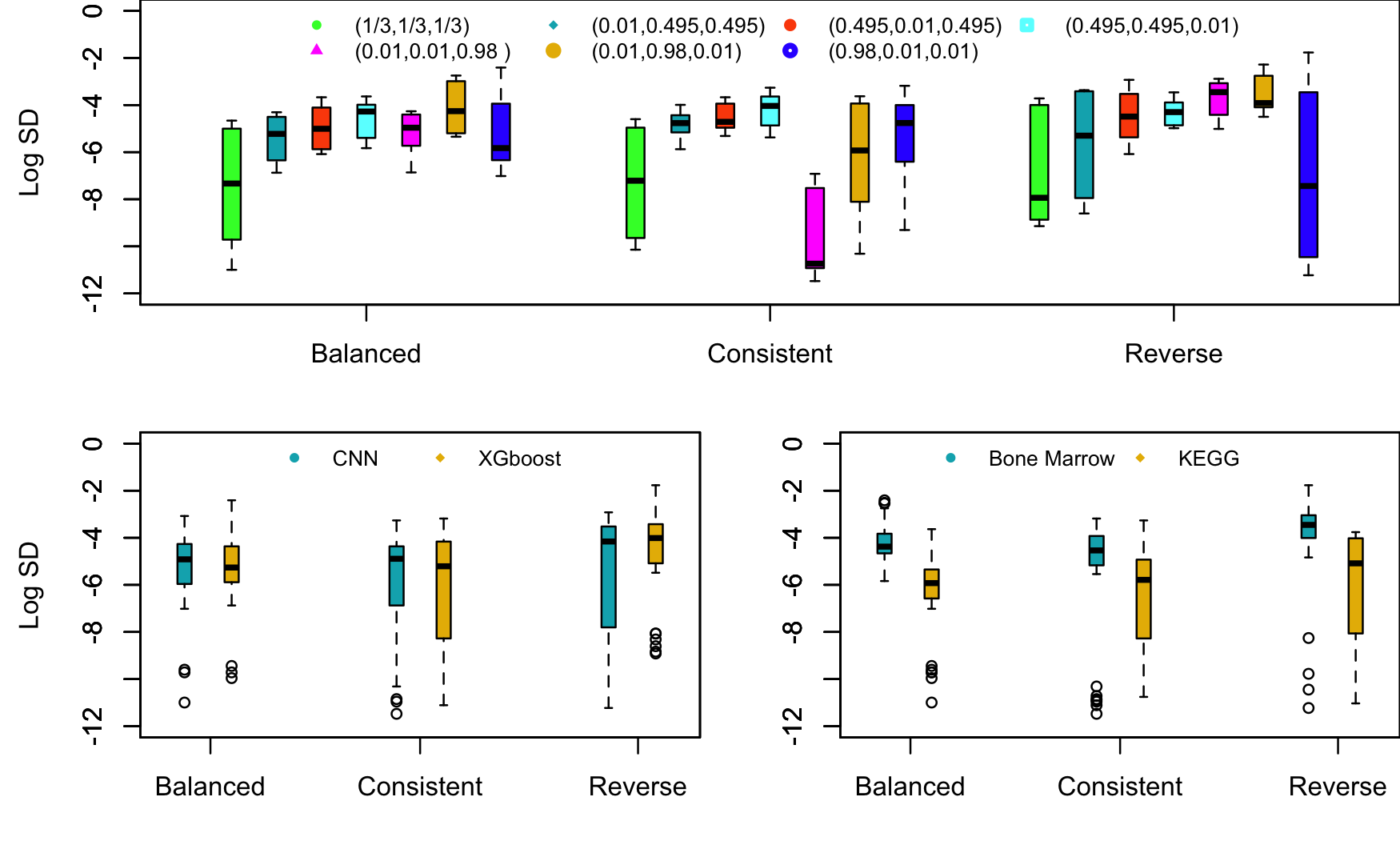}
\caption{Boxplots of the Log SD response under different levels of covariate variable and different combinations of label proportions for the three test scenarios.}\label{fig:log.s2}
\end{center}
\end{figure}

For the investigation of the Log SD as the response,
note that a small value of Log SD means small variation of AUC values of three classes.
That is, a smaller value of the Log SD is preferred.
For covariate $z_{1}$,  there is not a clear difference on Log SD between the CNN algorithm ($z_{1} = 0$) and the XGboost algorithm ($z_{1}=1$).
This implies that the two algorithms perform similarly with respect to the variation of AUC values from three classes.
Under the Balanced Scenario of the test dataset, it is seen that the Log SD is more stable (i.e., narrow range on the boxplots) than those under the other two scenarios
for covariate variable $z_{1}$.
But for covariate variable $z_{2}$, the Log SD under the KEGG dataset ($z_{2}=0$) is generally smaller than the Log SD under the Bone Marrow dataset ($z_{2} = 1$).
In contrast to $z_{2}$ being not that significant for the mean AUC,
the significance of $z_{2}$ for the Log SD implies that the KEGG dataset may provide more equally amount of information for three classes than the Bone Marrow dataset.


For the boxplots of the Log SD response under different combinations of label proportions,
the general pattern is similar to the pattern for the boxplots of the mean AUC response.
Specifically, the setting of balanced proportions of class label in the training dataset gives smaller Log SD than other settings.
It indicates that the balance among class labels in the training dataset could help algorithms to obtain a small variation of the AUC values of three classes.
When the proportions are dominated by two classes (e.g., $(x_{1}, x_{2}, x_{3}) = (0.01, 0.495, 0.495)$),
the setting gives slightly lower Log SD than the other two settings.
The interesting patterns are observed when there is only one class prevailing.
Under the Reverse Scenario, the range of Log SD for $(0.98,0.01,0.01)$ is the greatest, which may imply that such a setting could result in the most unstable classification performance.
Under the Consistent Scenario, the Log SD for the setting $(0.01,0.01,0.98)$ is smallest even compared to the settings under the Balanced Scenario.
One possible explanation is that when the class label 3 is dominating, the variation of AUC values for three classes becomes consistently low.


\subsection{Modeling Results}\label{sec: modleing-results}
In this section, we report analysis results of the regression model in Equation \eqref{eqn:reg.model} for the mean AUC and the Log SD as the response, respectively.
By fitting the regression model with the response of interest, we obtain estimated coefficients and their corresponding $t$-statistic and $p$-value.
The prediction performance is also investigated for different class label proportions.
The impact of predictor variables is assessed by the SHAP method.

\begin{table}
\begin{center}
\caption{Parameter estimation and testing statistics under the Balanced Scenario.}\label{tab: balanced-case}
\vspace{1.5ex}
\begin{tabular}{lrrrr|lrrrr}\hline\hline
\multicolumn{5}{c|}{Mean AUC Analysis} & \multicolumn{5}{c}{Log SD Analysis}\\\hline
Coef& Est& SE & $t$-value & $p$-value &
Coef& Est& SE & $t$-value & $p$-value\\\hline
$x_1$    & 0.4400   & 0.0173 & 25.440  & $<$0.001 &
$x_1$    & $-$4.631   & 0.565  & $-$8.202 & $<$0.001\\\hline
$x_2$    & 0.5455    & 0.0173  & 31.546 & $<$0.001 &
$x_2$    & $-$2.714    & 0.565  & $-$4.806 & $<$0.001\\\hline
$x_3$    & 0.8599   & 0.0173  & 49.764 & $<$0.001 &
$x_3$    & $-$3.937   & 0.565  & $-$6.997 & $<$0.001\\\hline
$x_1x_2$   & 0.5989   & 0.0513  & 11.627& $<$0.001 &
$x_1x_2$   & $-$3.228   & 1.677  & $-$1.919 & 0.059\\\hline
$x_1x_3$   & 0.6472   & 0.0513  & 12.604 & $<$0.001 &
$x_1x_3$   & $-$2.512   & 1.677  & $-$1.498 & 0.139\\\hline
$x_2x_3$   & 0.5512   & 0.0513  & 10.734 & $<$0.001 &
$x_2x_3$   & $-$6.725   & 1.677  & $-$4.011 & $<$0.001\\\hline
$x_1z_1$ & 0.2532 & 0.0195  & 12.975 & $<$0.001 &
$x_1z_1$ &  1.288 & 0.637  & 2.022 & 0.047 \\\hline
$x_2z_1$ & 0.1660 & 0.0195  & 8.506 & $<$0.001 &
$x_2z_1$ & $-$0.218 & 0.637  & $-$0.343 & 0.733 \\\hline
$x_3z_1$ & $-$0.0148 & 0.0195  & $-$0.758 & 0.451 &
$x_3z_1$ &  0.179 & 0.637  & 0.281 & 0.779\\\hline
$x_1z_2$ & 0.0241 & 0.0195  & 1.233 & 0.222 &
$x_1z_2$ & $-$1.727 & 0.637  & $-$2.711 & 0.008\\\hline
$x_2z_2$ & 0.0744 & 0.0195  & 3.815 & $<$0.001 &
$x_2z_2$ & $-$1.721 & 0.637  & $-$2.701 & 0.009\\\hline
$x_3z_2$ & $-$0.1186 & 0.0195 & $-$6.088 & $<$0.001 &
$x_3z_2$ & $-$1.850 & 0.637  & $-$2.909 & 0.005\\\hline
$z_1z_2$ & $-$0.0414 & 0.0160 & $-$2.593 & 0.012 &
$z_1z_2$ & $-$0.611 & 0.521 & $-$1.172 & 0.245\\ \hline\hline
\multicolumn{10}{c}{Implied effect for $z_1$ and $z_2$ }\\\hline
$z_{1}$ & 0.1348   &    0.0113 & 11.929   &  $<$0.001   &
$z_{1}$ &     0.416   &    0.369 & 1.127   &  0.263    \\ \hline
$z_{2}$ &     $-$0.007   &    0.0113 & $-$0.619   &  0.538 &
$z_{2}$ &     $-$1.766   &    0.369 & $-$4.785   &   $<$0.001  \\
\hline\hline
\end{tabular}
\end{center}
\end{table}

Table~\ref{tab: balanced-case} reports the analysis results for the Balanced Scenario.
For the mean AUC, among significant factors with $p$-values less than $0.05$,
the variable $x_{3}$ has the largest $t$ statistic.
It confirms our observation from the data visualization that the class label 3 play a significant role in the classification accuracy.
We also see that the estimated coefficients of interactions between two proportions (i.e. $x_{j}x_{j'}$'s) are all positive.
It implies that balance in the training dataset will increase the accuracy of the classification algorithms.
For covariate variable $z_{1}$ concerning two algorithms, the estimated coefficient of $z_1$ is positive with a statistically significant effect.
It confirms that the XGboost algorithm produces a higher mean AUC in comparison with the CNN algorithm.
For covariate variable $z_{2}$ concerning the two datasets (i.e.,  the KEGG and the Bone Marrow datasets),
the corresponding $p$-value is $0.538>0.05$, which confirms data visualization that $z_{2}$ does not provide significantly influence on the Mean AUC.

For the Log SD analysis, only the main effects of proportions and several interaction effects are statistically significant.
It is seen that covariate $z_{1}$ does not have a significant effect for the Log SD, which is consistent with the data visualization.
Note that $x_2x_3$ is significant with the largest $t$-statistic,
which may imply that the proportions of $x_2$ and $x_3$ are essential to minimize the dispersion of the AUC values among three classes.

\begin{table}
\begin{center}
\caption{Parameter estimation and testing statistics under the Consistent Scenario.}\label{tab:consistent-case}
\vspace{1.5ex}
\begin{tabular}{lrrrr|lrrrr}\hline\hline
\multicolumn{5}{c|}{Mean AUC Analysis} & \multicolumn{5}{c}{Log SD Analysis}\\\hline
Coef& Est& SE & $t$-value &$p$-value &
Coef& Est& SE & $t$-value &$p$-value\\\hline
$x_1$    & 0.4122   & 0.0273 & 17.386  & $<$0.001 &
$x_1$    & $-$3.649   & 0.836 & $-$4.367  & $<$0.001\\\hline
$x_2$  & 0.5833    & 0.0273 & 24.604  & $<$0.001 &
$x_2$  & $-$5.523    & 0.836 & $-$6.608   & $<$0.001\\\hline
$x_3$   & 1.0009   & 0.0273 & 42.251  & $<$0.001 &
$x_3$   & $-$10.448   & 0.836 & $-$12.511 & $<$0.001\\\hline
$x_1x_2$   & 0.5091   & 0.0704 & 7.210  & $<$0.001 &
$x_1x_2$   & 1.514   & 2.481  & 0.608 & 0.545\\\hline
$x_1x_3$   & 0.6150   & 0.0704 & 8.737  & $<$0.001 &
$x_1x_3$   & 7.079   & 2.481 & 2.853 & 0.006\\\hline
$x_2x_3$   & 0.3838   & 0.0704 & 5.453  & $<$0.001 &
$x_2x_3$   & 7.823   & 2.481 & 3.153 & 0.002\\\hline
$x_1z_1$ & 0.3068 & 0.0267 & 11.471  & $<$0.001 &
$x_1z_1$ & $-$0.102 & 0.943 & $-$0.109 & 0.914\\\hline
$x_2z_1$ & 0.1576 & 0.0267 & 5.894  & $<$0.001 &
$x_2z_1$ & 1.185 & 0.943  & 1.257 & 0.213\\\hline
$x_3z_1$ & $-$0.0366 & 0.0267 & $-$1.371  & 0.175 &
$x_3z_1$ & 0.428 & 0.943 &0.455 & 0.651\\\hline
$x_1z_2$ & 0.0290 & 0.0267 & 1.083  & 0.282 &
$x_1z_2$ & $-$1.922 & 0.943 & $-$2.038 & 0.045\\\hline
$x_2z_2$ & 0.1085 & 0.0267  & 4.058 & $<$0.001 &
$x_2z_2$ & $-$1.283 & 0.943  & $-1.361$ & 0.178\\\hline
$x_3z_2$ & $-$0.1943 & 0.0267 & $-$7.273 & $<$0.001 &
$x_3z_2$ & 2.498 & 0.943  & 2.653 & 0.010\\\hline
$z_1z_2$ & $-$0.0189 & 0.0219 & $-$0.861  & 0.392 &
$z_1z_2$ & $-$2.002 & 0.772 & $-$2.594 & 0.012\\\hline\hline
\multicolumn{10}{c}{Implied effect for $z_1$ and $z_2$ }\\\hline
$z_{1}$ & 0.1426   &    0.0155 & 9.200  &  $<$0.001   &
$z_{1}$ &     0.504   &    0.546 & 0.923   &  0.359   \\\hline
$z_{2}$ &     $-$0.0189   &    0.0155 & $-$1.219   &  0.227   &
$z_{2}$ &     $-$0.236   &    0.546   & $-$0.432   & 0.667  \\\hline\hline
\end{tabular}
\end{center}
\end{table}

Table~\ref{tab:consistent-case} reports the analysis results for the Consistent Scenario.
It is seen that almost all terms involving the class proportion $x_{3}$ are significant for both mean AUC and Log SD.
This implies that the proportion of class label 3 is not only of most importance to maintain the classification accuracy,
but also important for the variation of the AUC values of the three classes.
The covariate variable $z_1$ is significant with a positive coefficient,
indicating that the XGboost algorithm performs better than the CNN algorithm for the mean AUC in the Consistent Scenario.
For the Log SD, the estimated coefficient for $x_1x_3$, which is not significant in the Balanced Scenario, becomes statistically significant.
A possible explanation is that the proportion of label 1 becomes more important in the Consistent Scenario than that in the Balanced Scenario.
Under the Consistent Scenario, the results in Table~\ref{tab:consistent-case} show that both covariates $z_1$ and $z_2$ seem not having much influence on the Log SD.

\begin{table}
\begin{center}
\caption{Parameter estimation and testing statistics under the Reverse Scenario.}\label{tab:reverse-case}
\vspace{1.5ex}
\begin{tabular}{lrrrr|lrrrr}\hline\hline
\multicolumn{5}{c|}{Mean AUC Analysis} & \multicolumn{5}{c}{Log SD Analysis}\\\hline
Coef & Est& SE & $t$-value &$p$-value &
Coef& Est& SE & $t$-value &$p$-value\\\hline
$x_1$    & 0.4755   & 0.0205 &23.245  & $<$0.001 &
$x_1$    & $-$8.713   & 0.716  & $-$12.162 & $<$0.001\\\hline
$x_2$  & 0.5211    & 0.0205 &25.472  & $<$0.001 &
$x_2$  & $-$2.181    & 0.716 & $-$3.044 & 0.003 \\\hline
$x_3$   & 0.7350   & 0.0205 &35.960  & $<$0.001 &
$x_3$   & $-$1.873   & 0.716 &$-$2.617 & 0.011\\\hline
$x_1x_2$   & 0.6454   & 0.0607  &10.594 & $<$0.001 &
$x_1x_2$   & 1.594   & 2.127 & 0.747 & 0.458\\\hline
$x_1x_3$   & 0.6990   & 0.0607  &11.509 & $<$0.001 &
$x_1x_3$   & 1.303   & 2.127 &0.613 & 0.542\\\hline
$x_2x_3$   & 0.6472   & 0.0607 & 10.656   & $<$0.001 &
$x_2x_3$   & $-$11.262   & 2.127 &$-$5.295 & $<$0.001\\\hline
$x_1z_1$ & 0.1807 & 0.0231  &7.831 & $<$0.001 &
$x_1z_1$ & 4.950 & 0.808 & 6.125 &  $<$0.001\\\hline
$x_2z_1$ & 0.1737 & 0.0231 &7.524  & $<$0.001 &
$x_2z_1$ & $-$0.463 & 0.808 &$-$0.573 & 0.568\\\hline
$x_3z_1$ & $-$0.0208 & 0.0231 & $-$0.901  & 0.371 &
$x_3z_1$ & $-$0.233 & 0.808  &$-$0.288 & 0.774\\\hline
$x_1z_2$ & 0.0105 & 0.0231  &0.456 & 0.650 &
$x_1z_2$ & $-$1.132 & 0.808 &$-$1.400 & 0.166\\\hline
$x_2z_2$ & 0.0150 & 0.0231  &0.650 & 0.517 &
$x_2z_2$ & $-$1.449 & 0.808 &$-$1.792 & 0.077\\\hline
$x_3z_2$ & $-$0.0576 & 0.0231  & $-$2.500  & 0.015 &
$x_3z_2$ & $-$2.635 & 0.808  & $-$3.265 & 0.002\\\hline
$z_1z_2$ & $-$0.0319 & 0.0189 & $-$1.688 &  0.096 &
$z_1z_2$ & $-$0.695 & 0.661  &$-$1.051 & 0.297\\\hline\hline
\multicolumn{10}{c}{Implied effect for $z_1$ and $z_2$ }\\\hline
$z_{1}$ & 0.1112   &    0.0114 &9.754   &  $<$0.001   &
$z_{1}$ &     1.418   &    0.468 & 3.030   &  0.003   \\\hline
$z_{2}$ &     $-$0.012   &    0.0114 & $-$1.053   &  0.296  &
$z_{2}$ &     $-$1.739   &    0.468   & $-$3.716 &  $<$0.001 \\\hline\hline
\end{tabular}
\end{center}
\end{table}

Table~\ref{tab:reverse-case} reports the analysis results for the Reverse Scenario.
The significance of $x_{3}$ (i.e., the proportion of class 3) has the similar pattern as what we observed in both the balanced scenario and the consistent scenario.
The covariate $z_{1}$ is significant for the mean AUC, but it is not significant for the Log SD.
In contrast, covariate $z_{2}$ is significant for the log SD, but it is not significant for the mean AUC.
It is interesting to note that, under the Reverse Scenario,  the $t$-statistic for $x_{1}$ is the largest in absolute value for the Log SD.
It may indicate that the proportion of label 1 becomes crucial to affect the variation of AUC values among three classes in the Reverse Scenario.

\begin{figure}
\centering
\begin{tabular}{ccc}
\includegraphics[width=.3\textwidth]{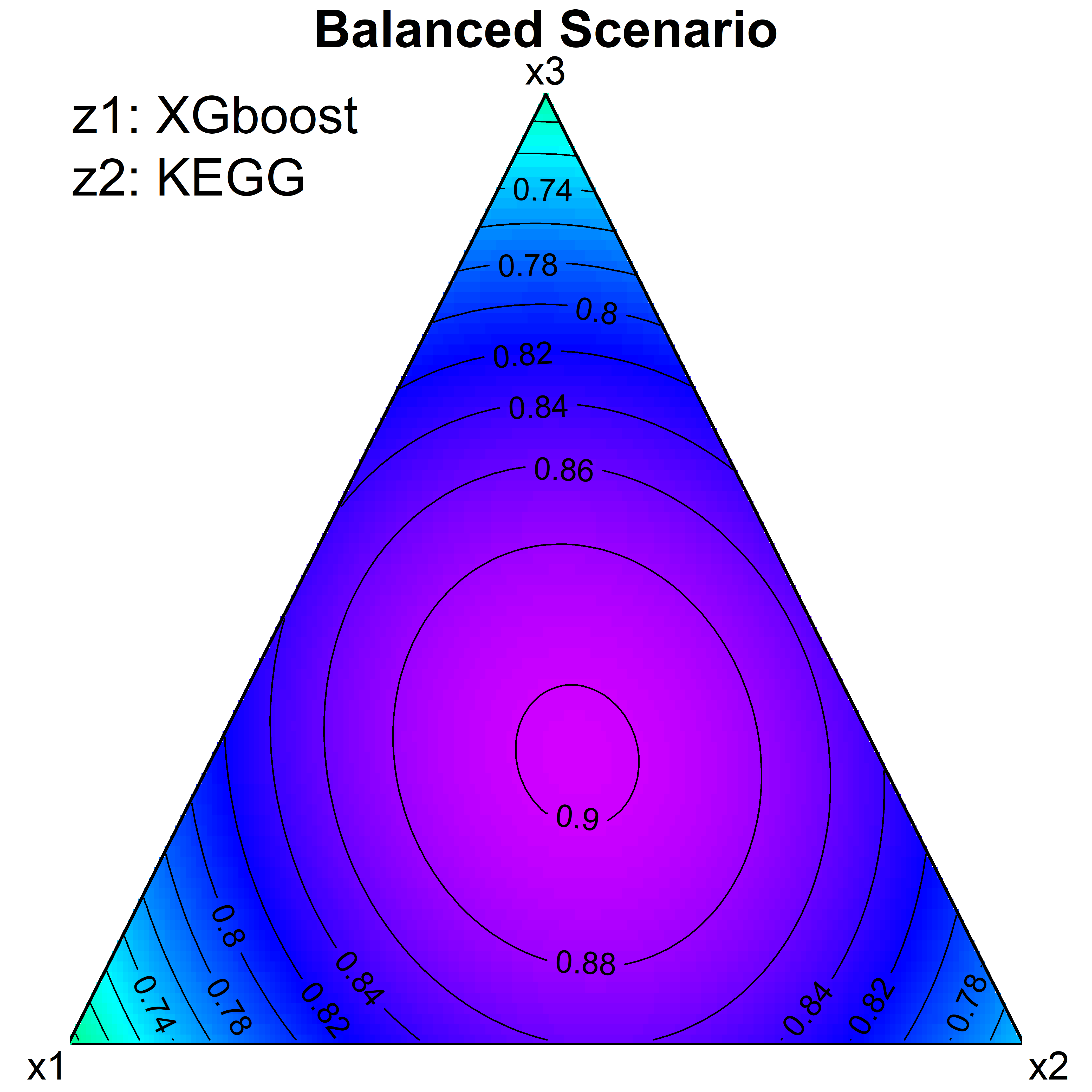}&
\includegraphics[width=.3\textwidth]{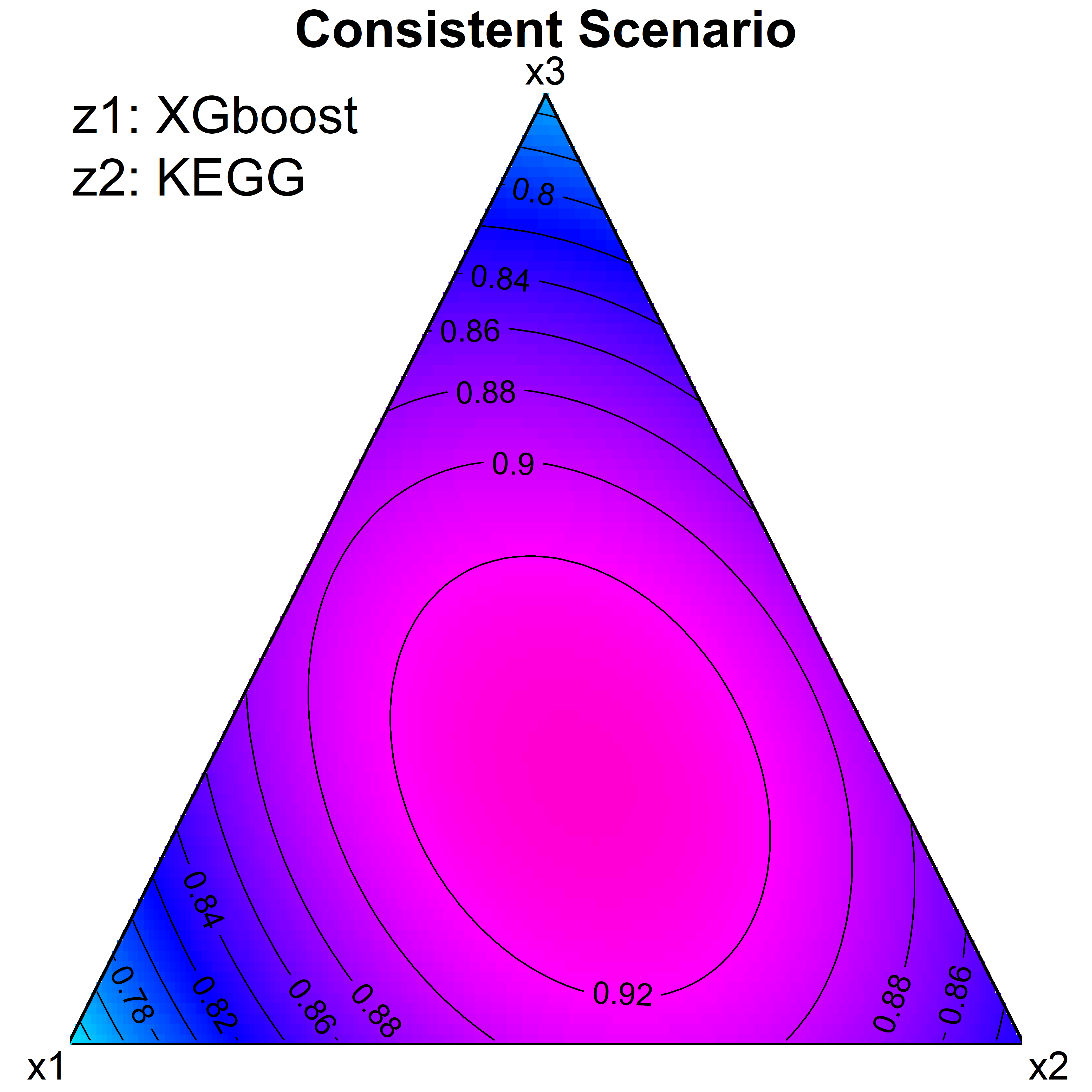}&
\includegraphics[width=.3\textwidth]{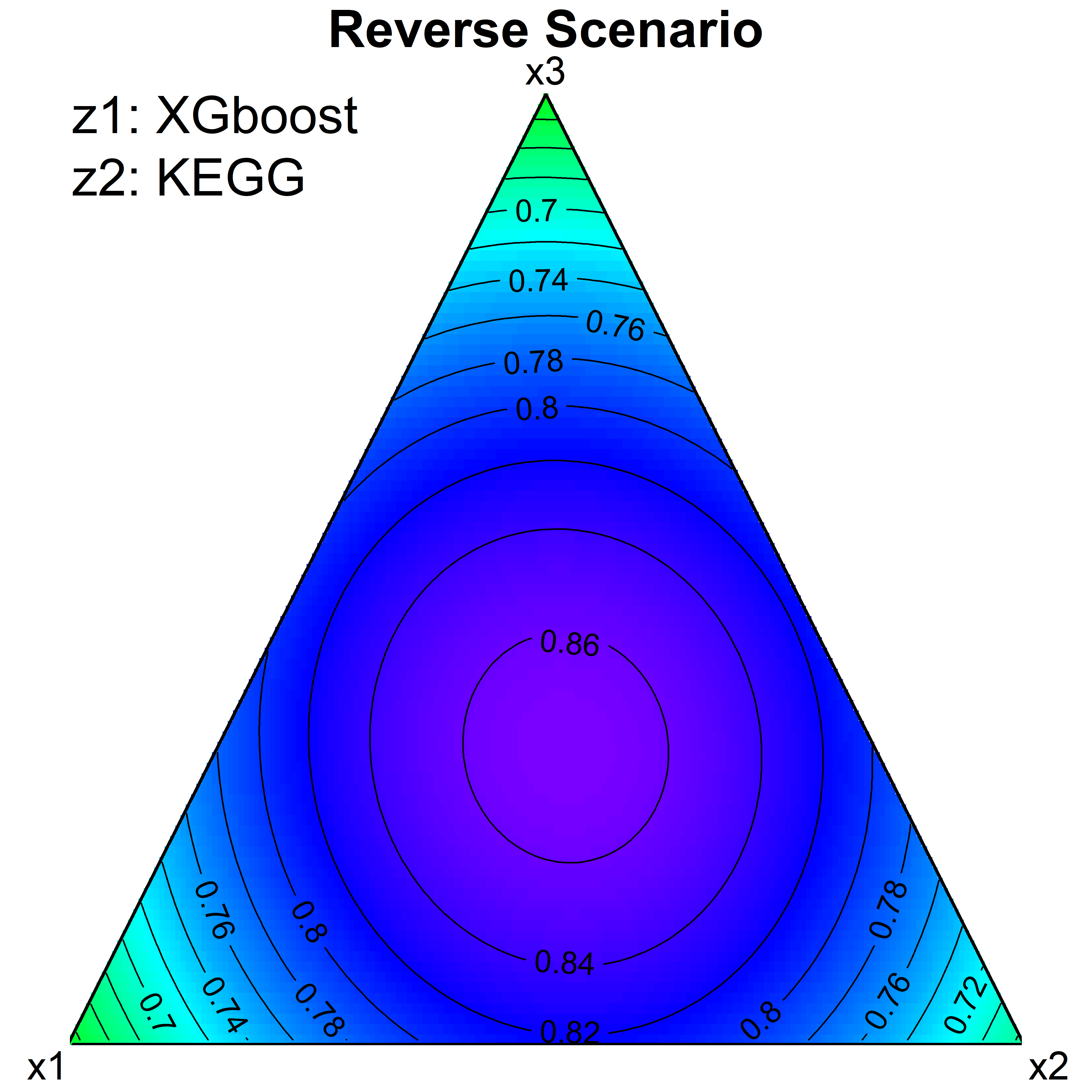}\\
\includegraphics[width=.3\textwidth]{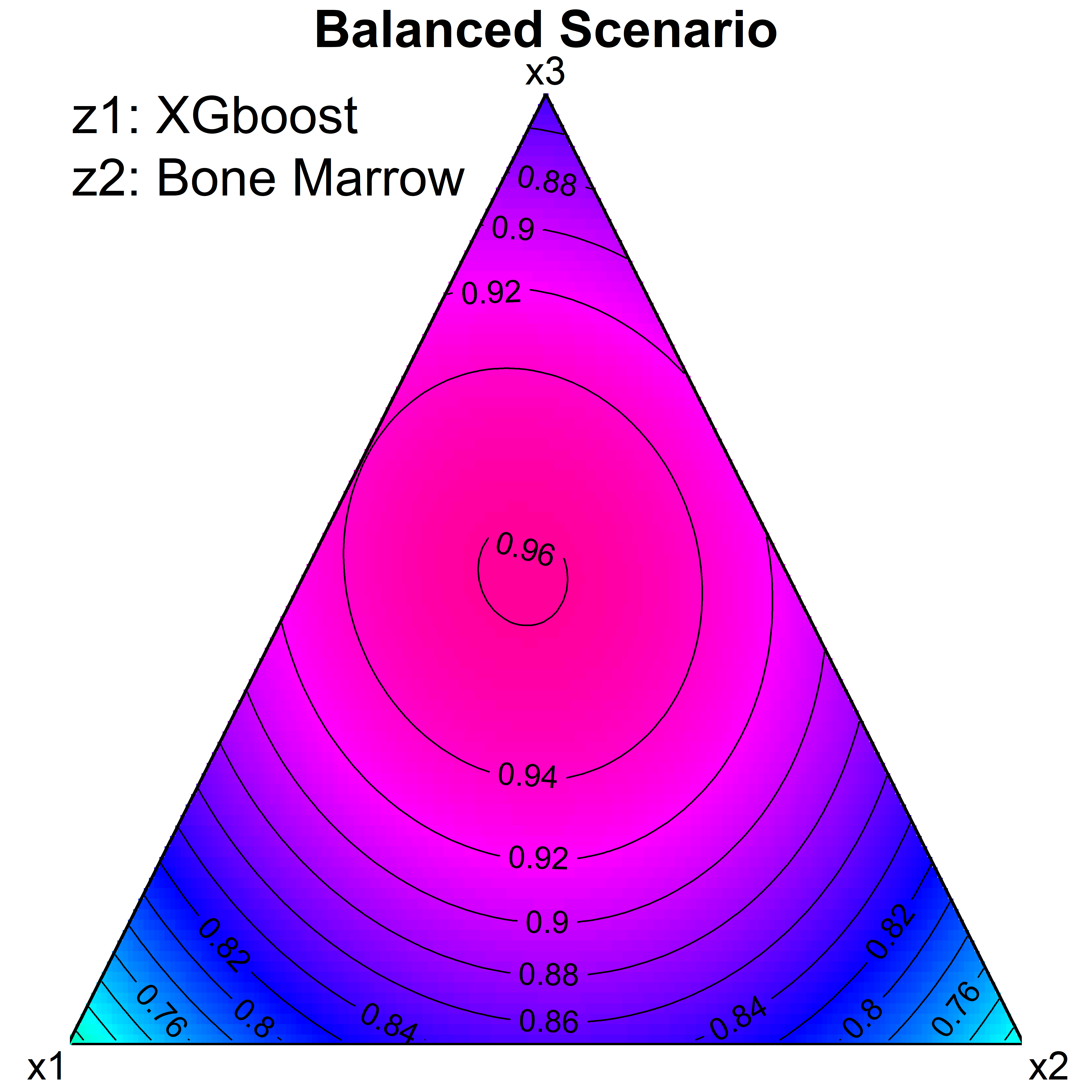}&
\includegraphics[width=.3\textwidth]{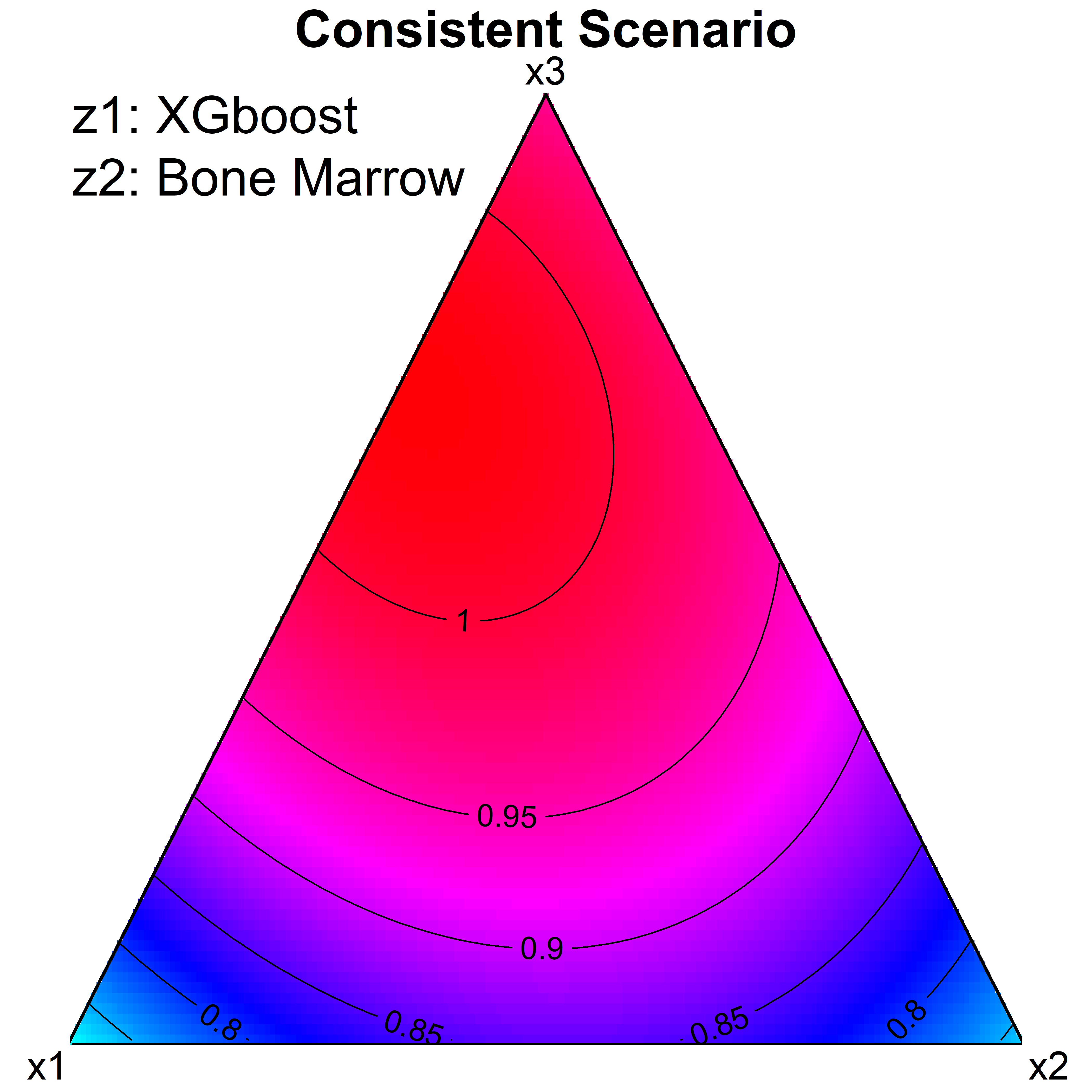}&
\includegraphics[width=.3\textwidth]{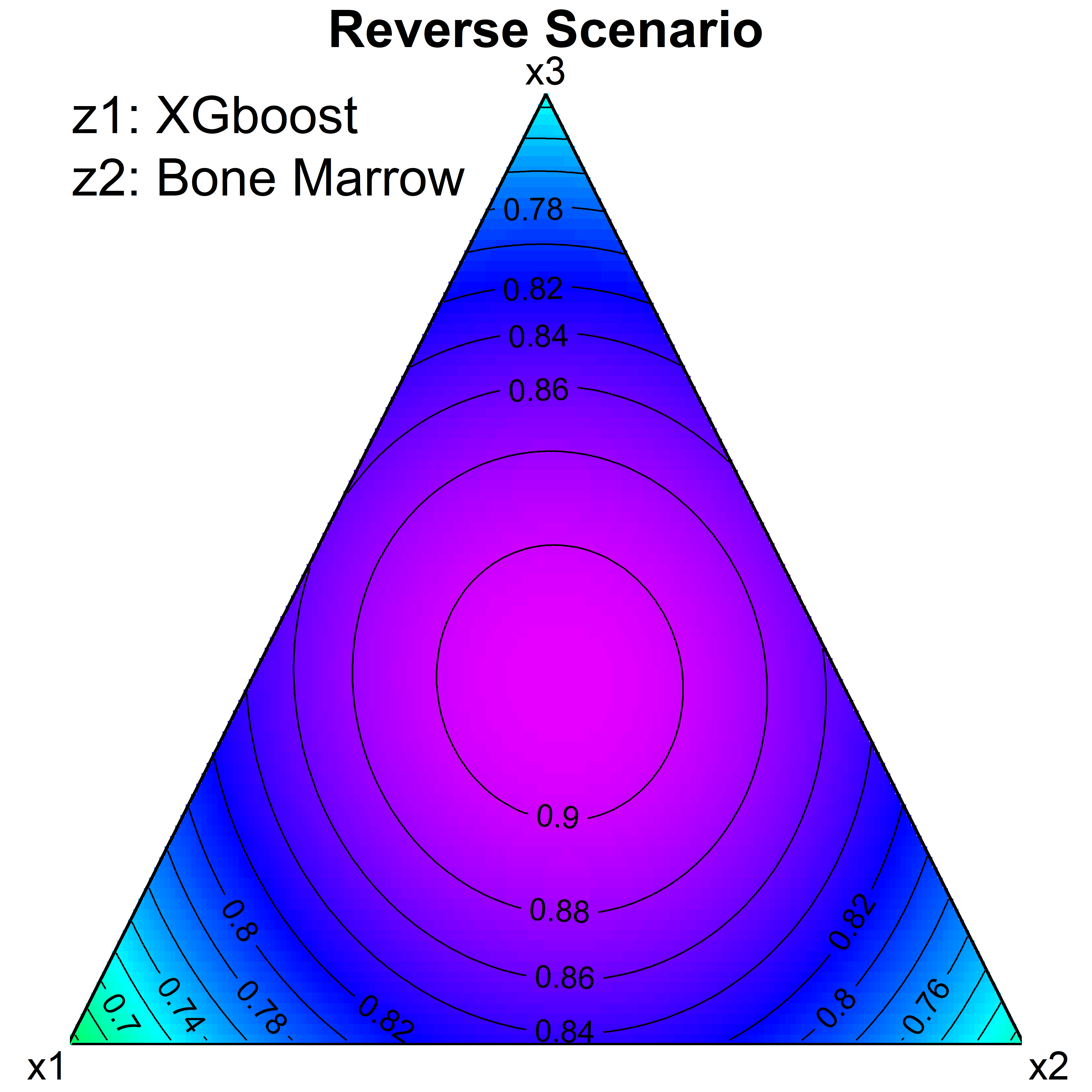}\\
\includegraphics[width=.3\textwidth]{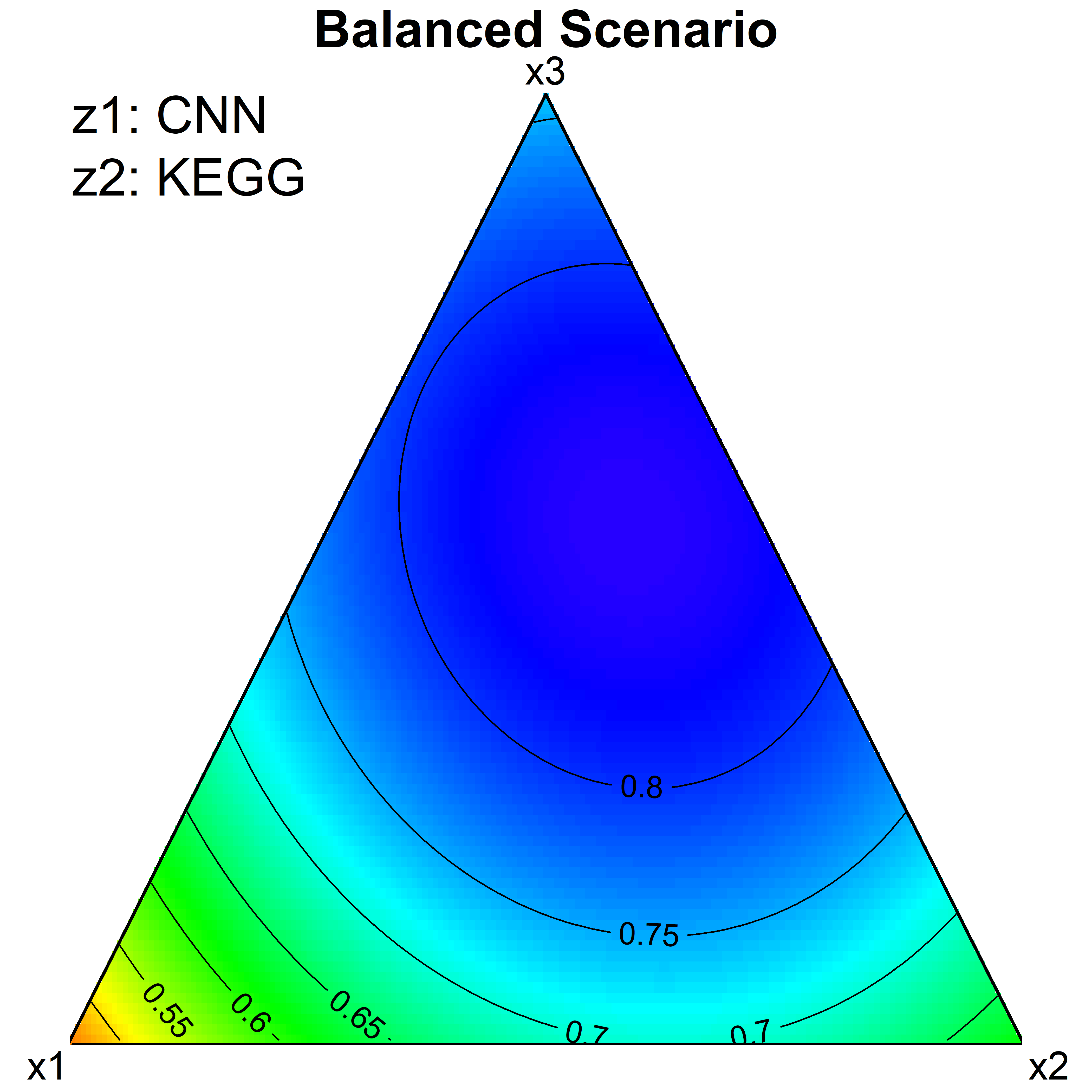}&
\includegraphics[width=.3\textwidth]{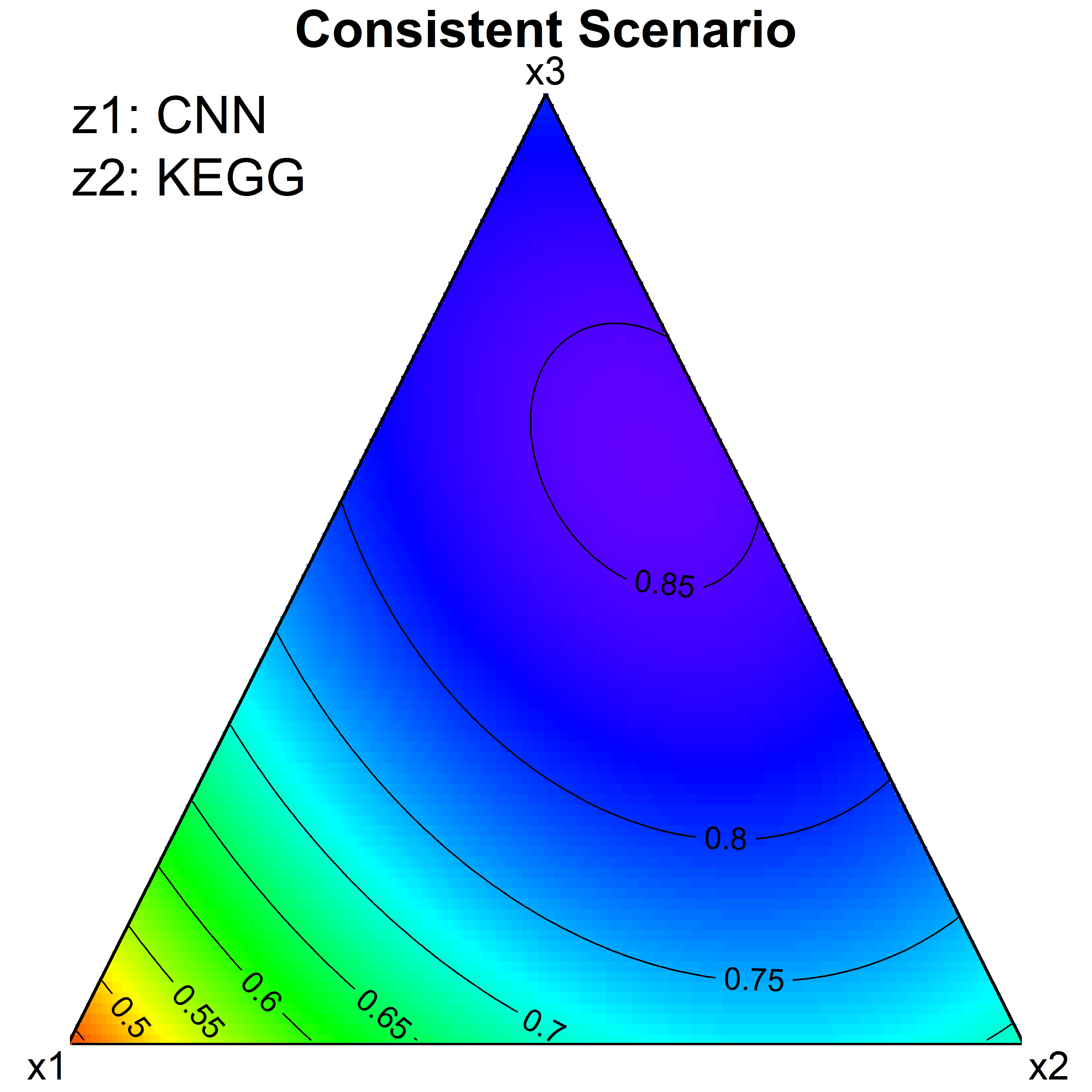}&
\includegraphics[width=.3\textwidth]{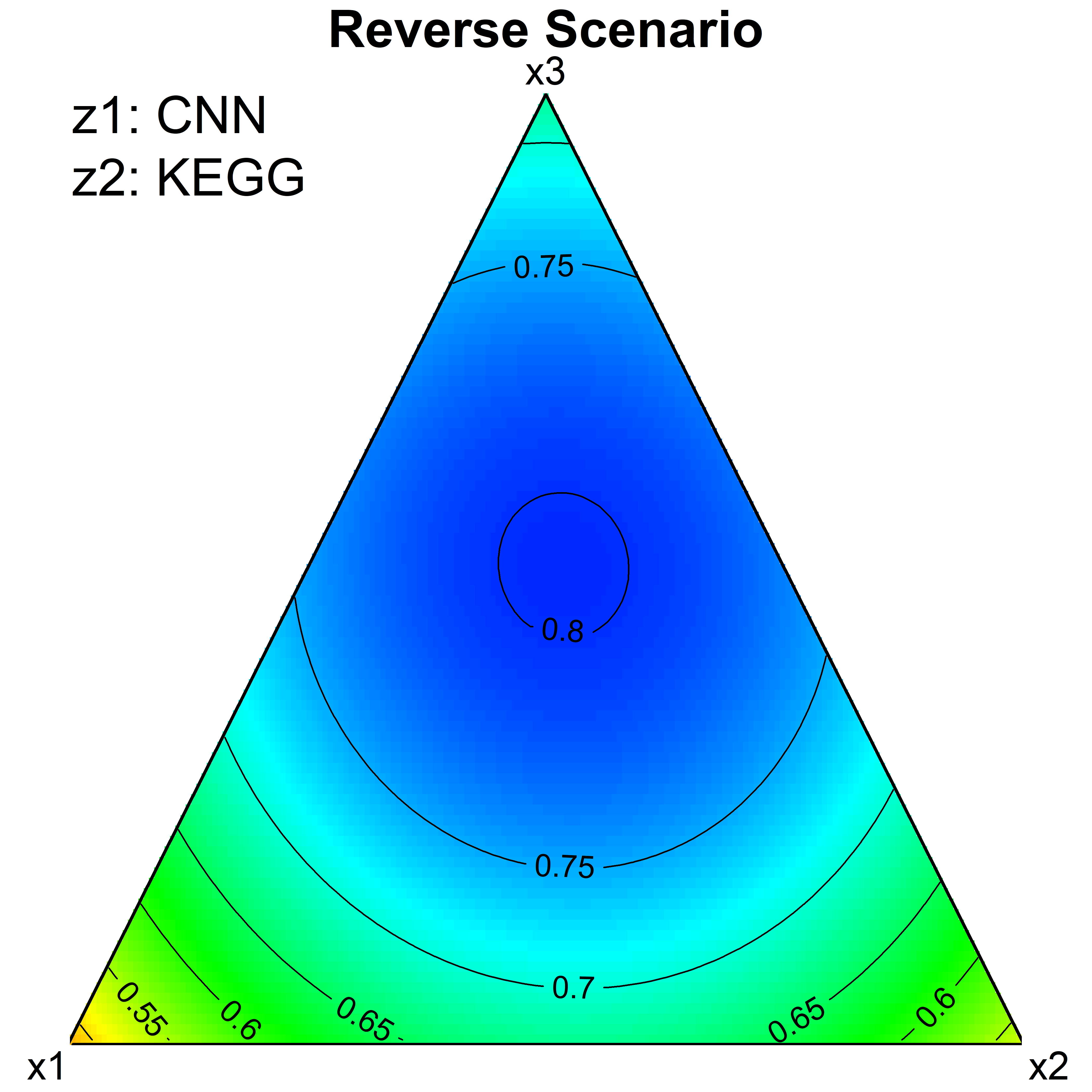}\\
\includegraphics[width=.3\textwidth]{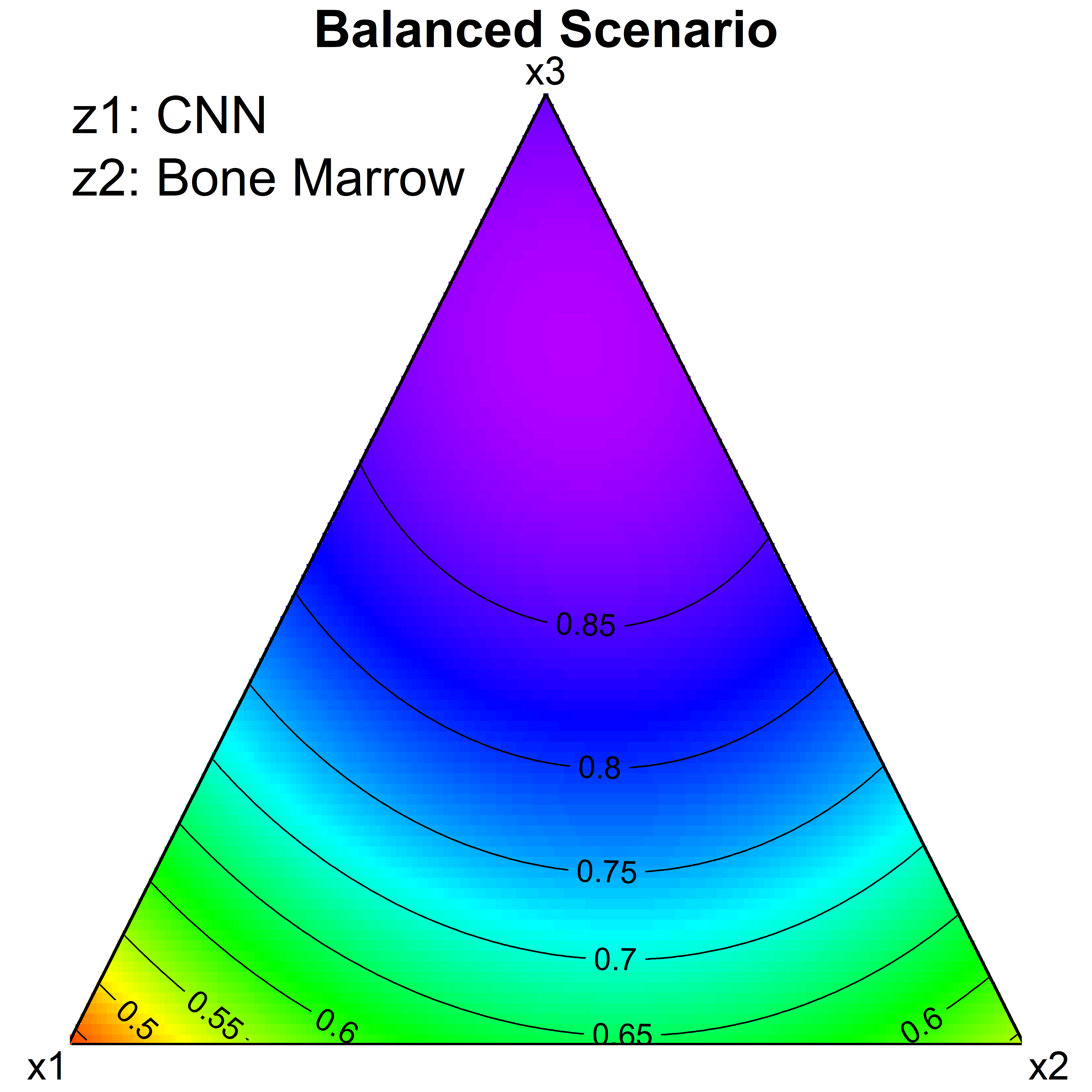}&
\includegraphics[width=.3\textwidth]{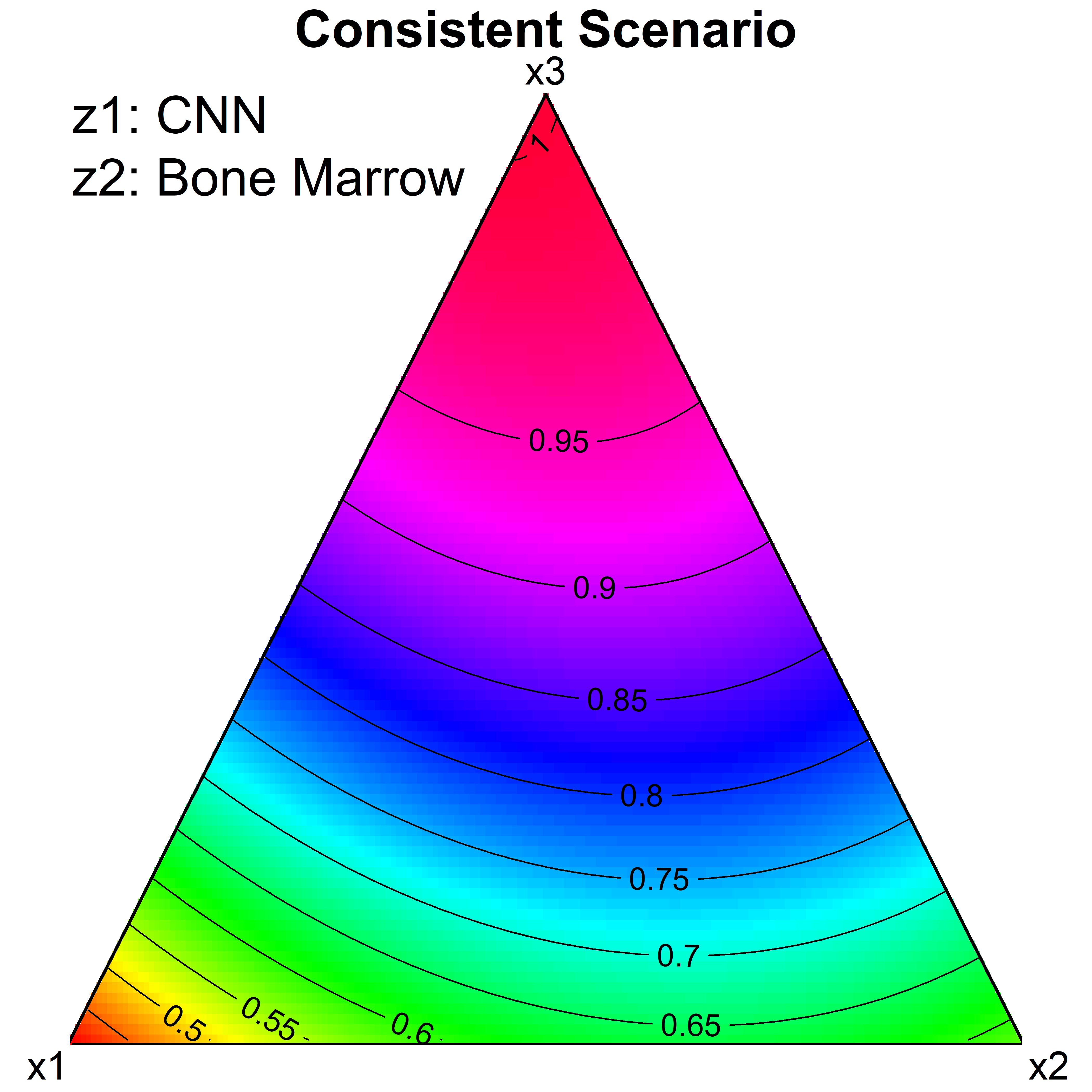}&
\includegraphics[width=.3\textwidth]{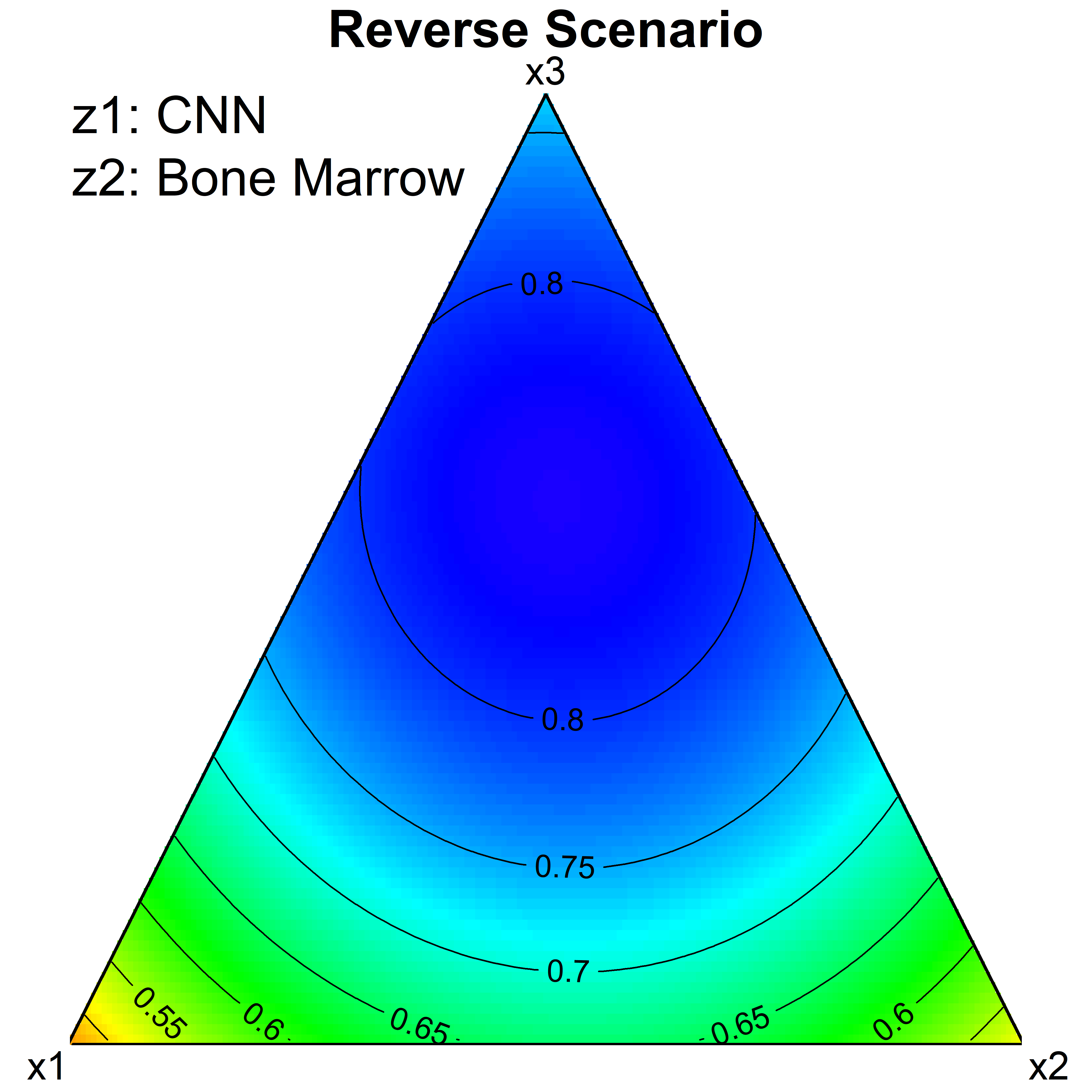}\\
\end{tabular}
\caption{The triangle contour plots of prediction for the Mean AUC under three scenarios.}\label{fig:contour-AUC}
\end{figure}

Furthermore, we evaluate the performance of the estimated model at different proportions of class labels $(x_{1}, x_{2}, x_{3})$.
Specifically, given a level combination of covariate variables $z_{1}$ and $z_{2}$, we provide the triangle contour plots of prediction at $(x_{1}, x_{2}, x_{3})$.
Figure \ref{fig:contour-AUC} shows the triangle contour plots of prediction for the Mean AUC under different level combinations of the covariate variables.
Based those results in Figure \ref{fig:contour-AUC}, the prediction accuracy of the classification algorithms (i.e., the predicted mean AUC) generally achieves the best performance when the class proportions are balanced (i.e., $(x_{1}, x_{2}, x_{3} = (1/3, 1/3, 1/3)$).
However, comparing the XGboost ($z_{1} = 1$) and the CNN ($z_{1} = 0$), the XGboost algorithm appears to be more symmetric with respect to three vertex $x_{1}$, $x_{2}$, and $x_{3}$.
While the prediction accuracy of the CNN algorithm appears to be more affected by the proportion of the class 3.
It is noted that, under the setting of the Consistent Scenario,
both classification algorithms can achieve the perfect accuracy (i.e., prediction of the mean AUC being 1) under the setting of class proportion $x_{3}$ prevailing.
One plausible explanation is that data points from class 3 are more important than the data points from the other two classes in terms of affecting the classification accuracy for the KEGG and the Bone Marrow datasets in this study.

Figure~\ref{fig:contour-AUC} shows the triangle contour plots of prediction for the Log SD under different level combinations of the covariate variables.
The patterns of the contour plots are generally more heterogeneous across different level combinations of covariate variables $z_{1}$ and $z_{2}$.
One interesting observation is that the prediction pattern in the contour plot is not symmetric with respect to three vertex $x_{1}$, $x_{2}$, and $x_{3}$.
It may imply that the data points from the three classes do not equally contribute to the classification performance in terms of the variation of the AUC values from three classes.

\begin{figure}
\centering
\begin{tabular}{ccc}
\includegraphics[width=.3\textwidth]{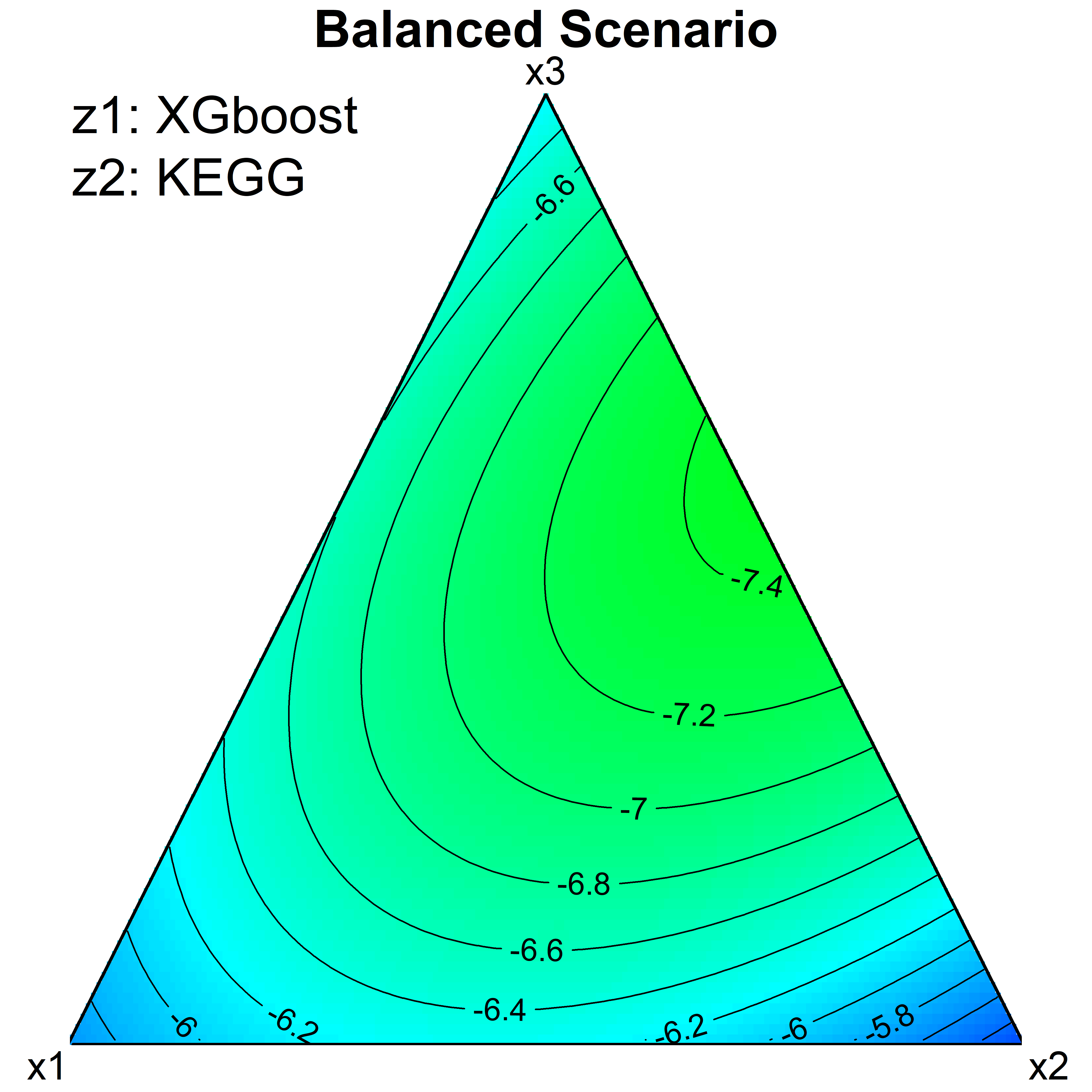}&
\includegraphics[width=.3\textwidth]{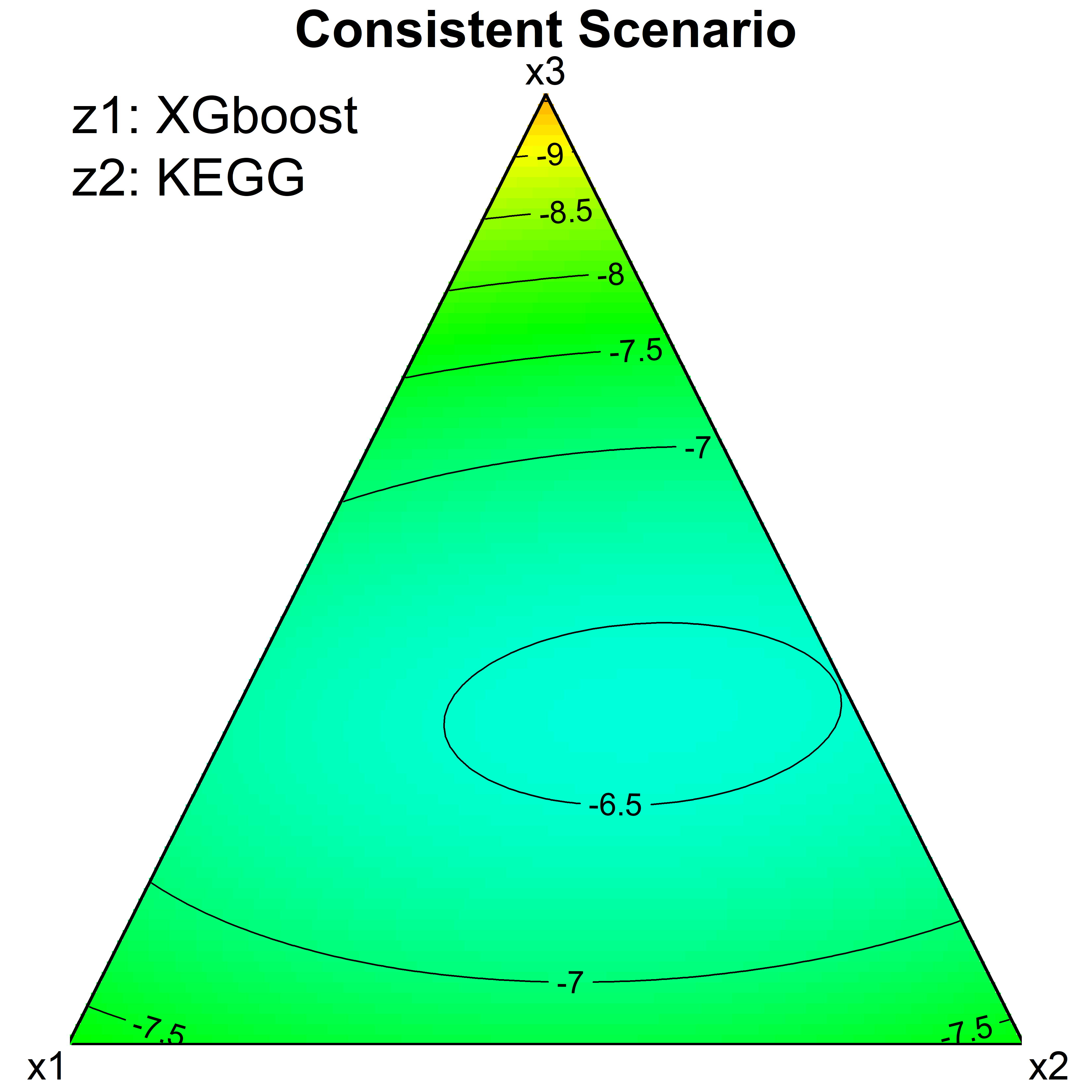}&
\includegraphics[width=.3\textwidth]{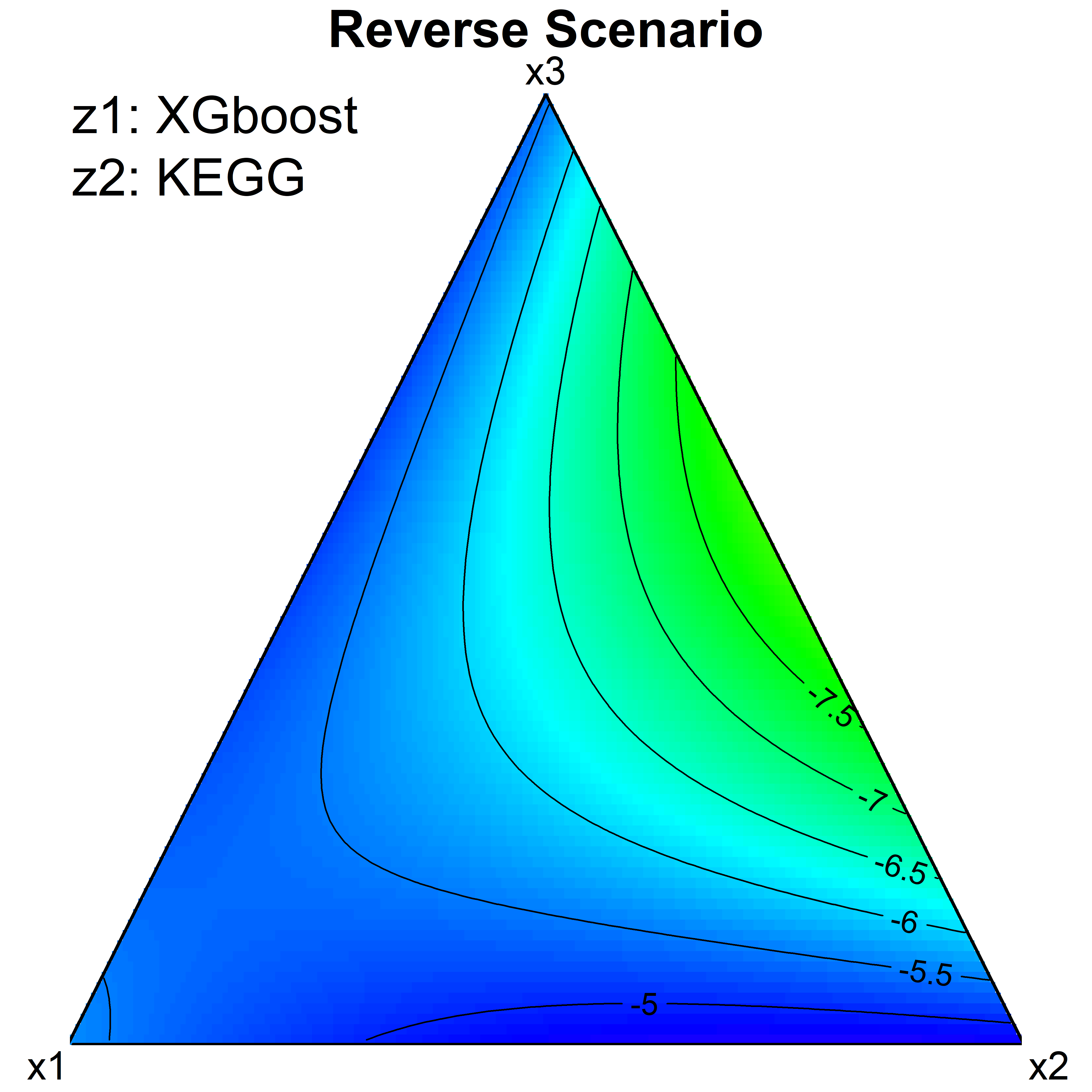}\\
\includegraphics[width=.3\textwidth]{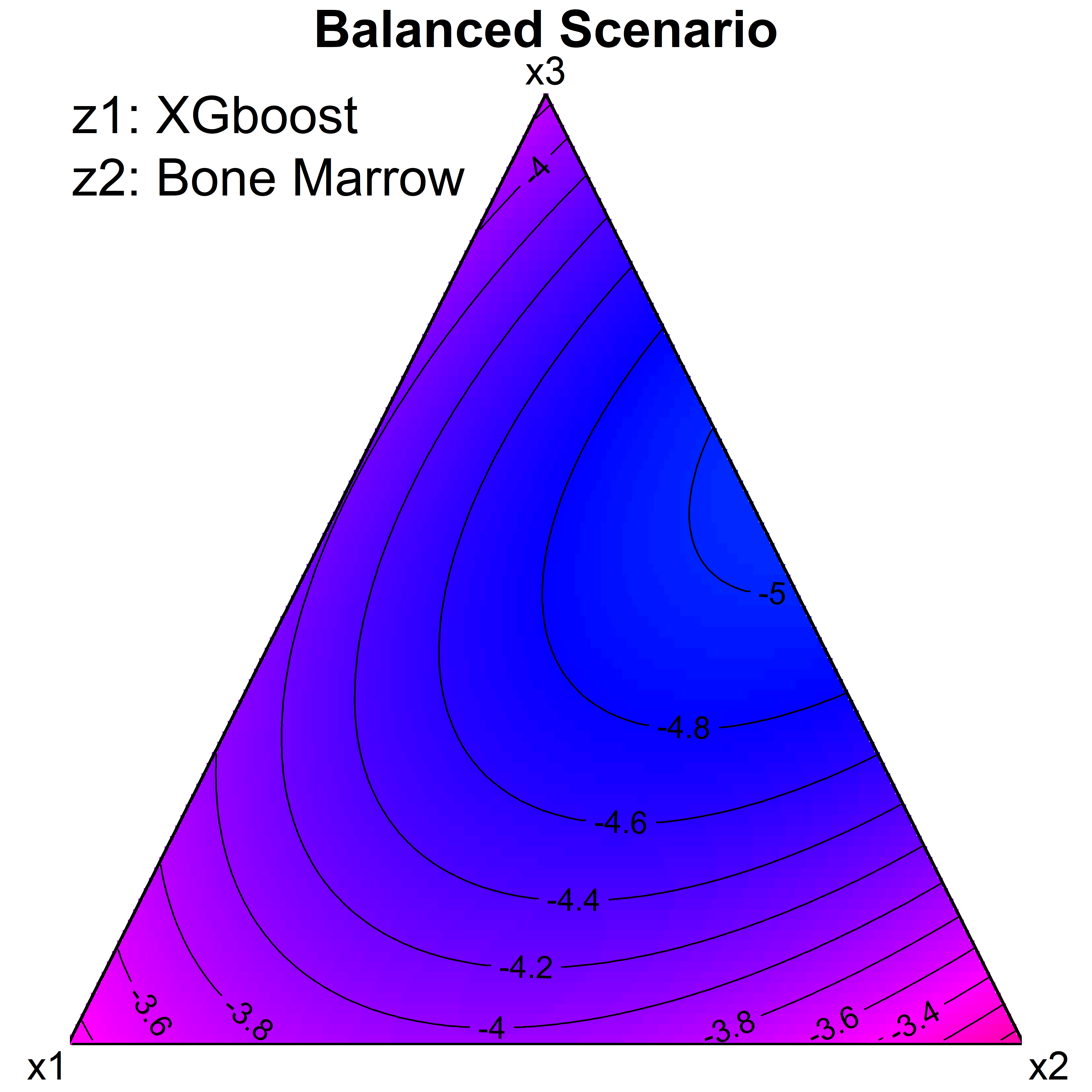}&
\includegraphics[width=.3\textwidth]{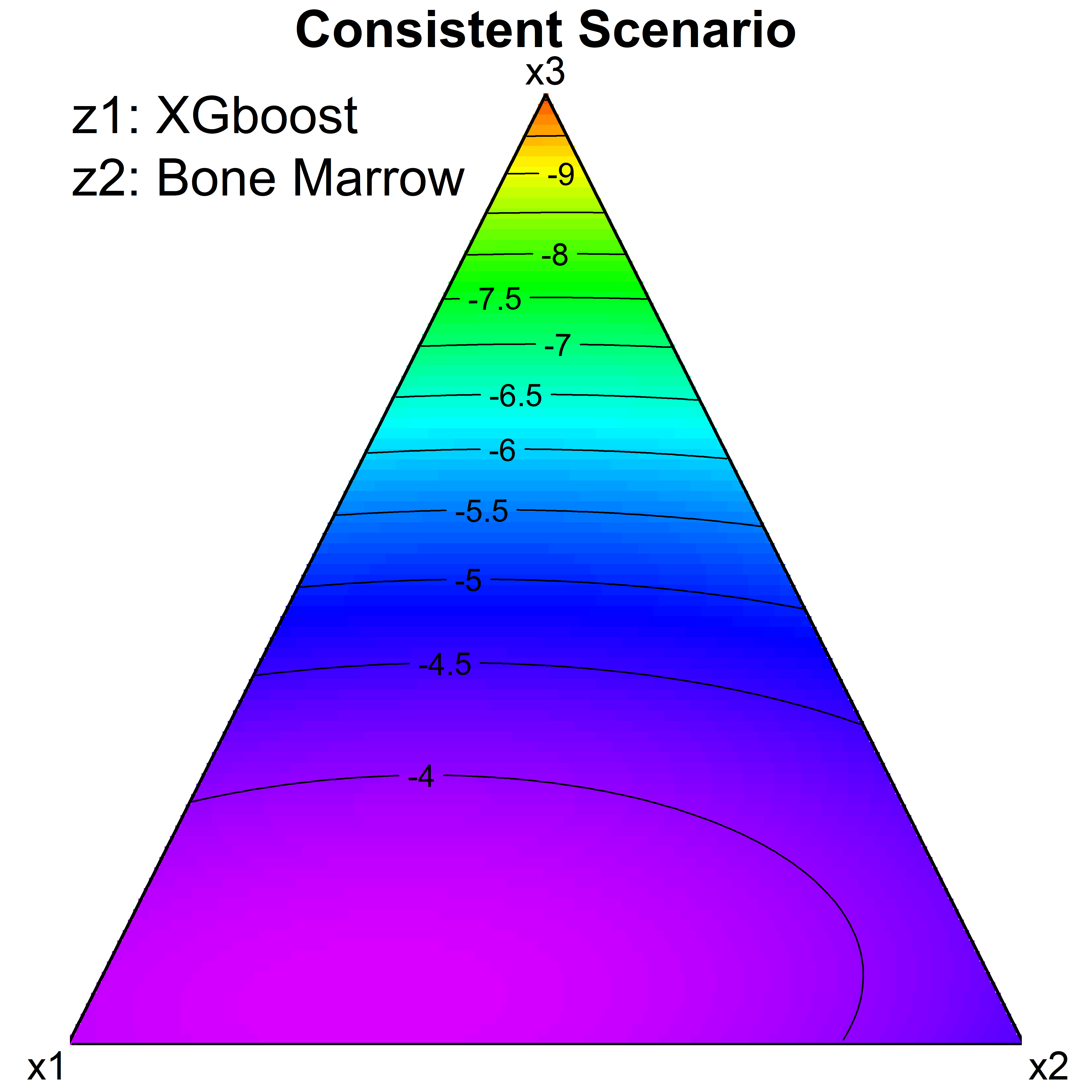}&
\includegraphics[width=.3\textwidth]{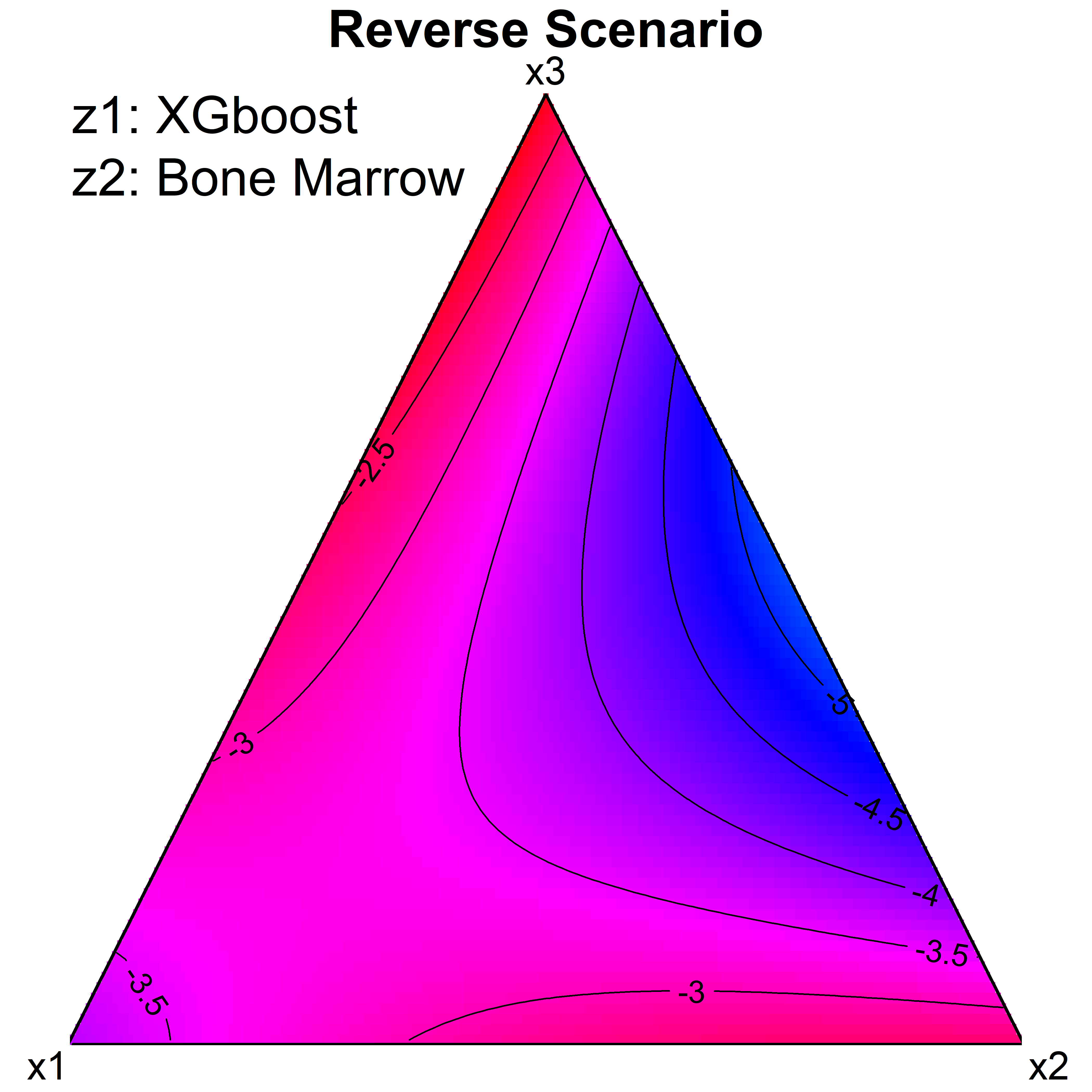}\\
\includegraphics[width=.3\textwidth]{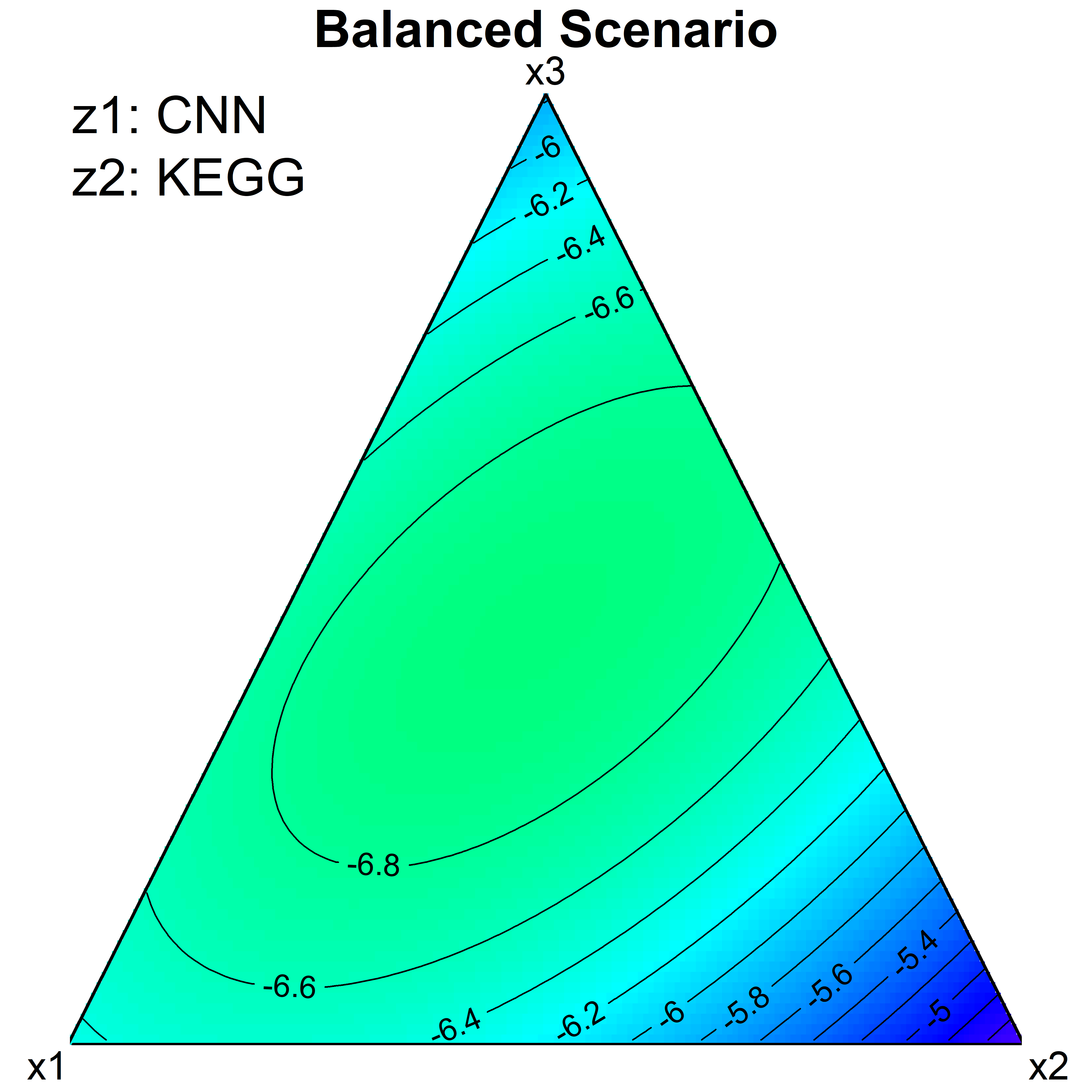}&
\includegraphics[width=.3\textwidth]{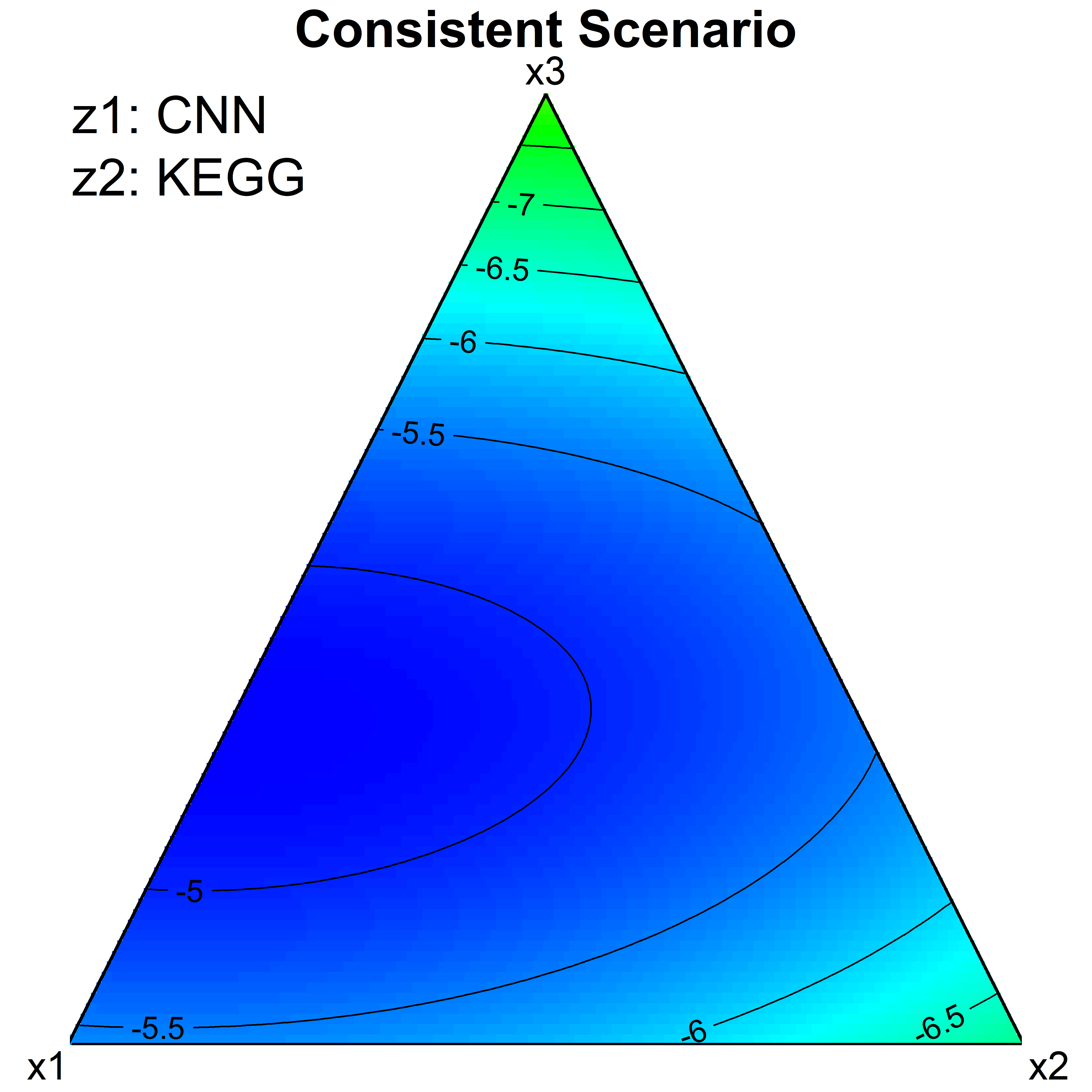}&
\includegraphics[width=.3\textwidth]{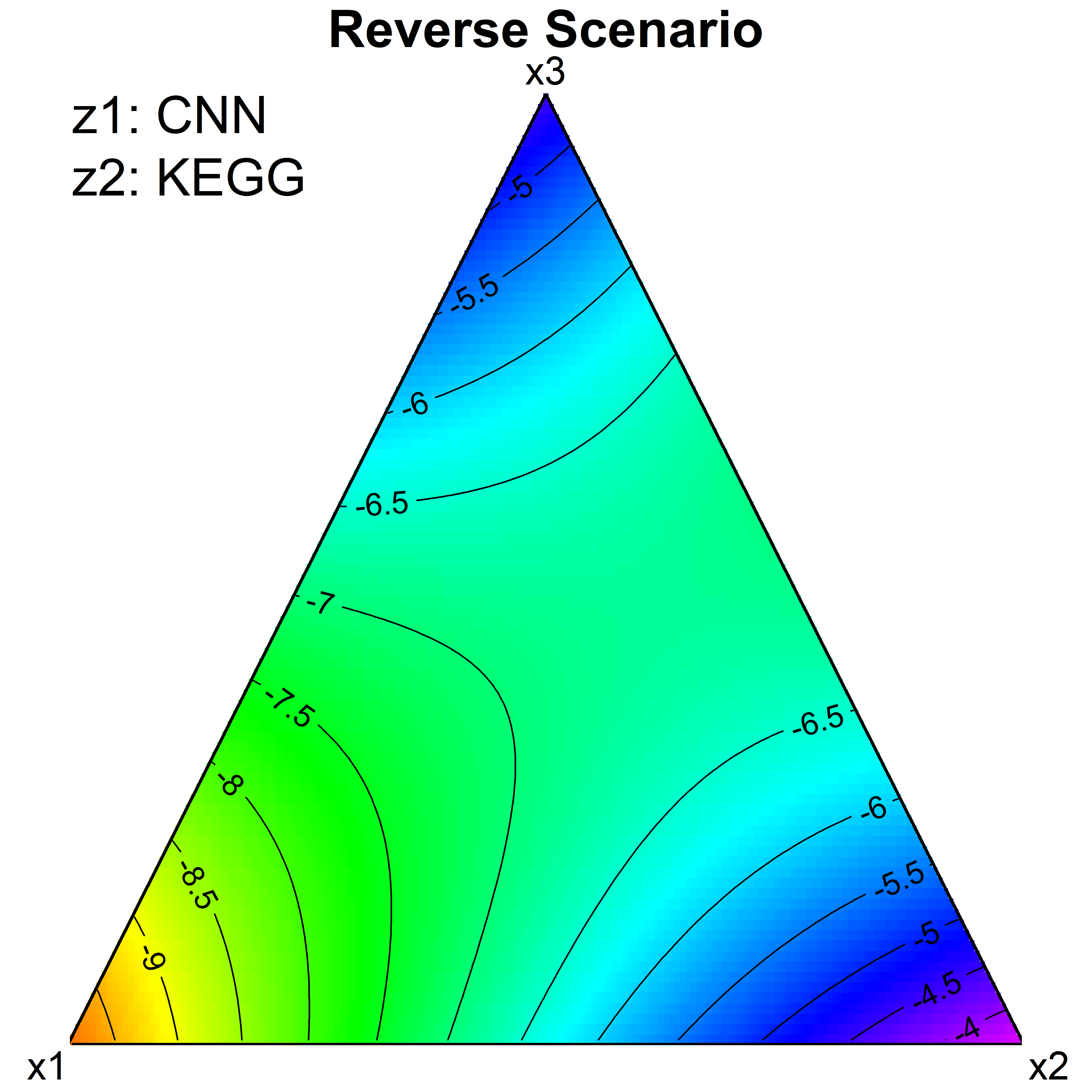}\\
\includegraphics[width=.3\textwidth]{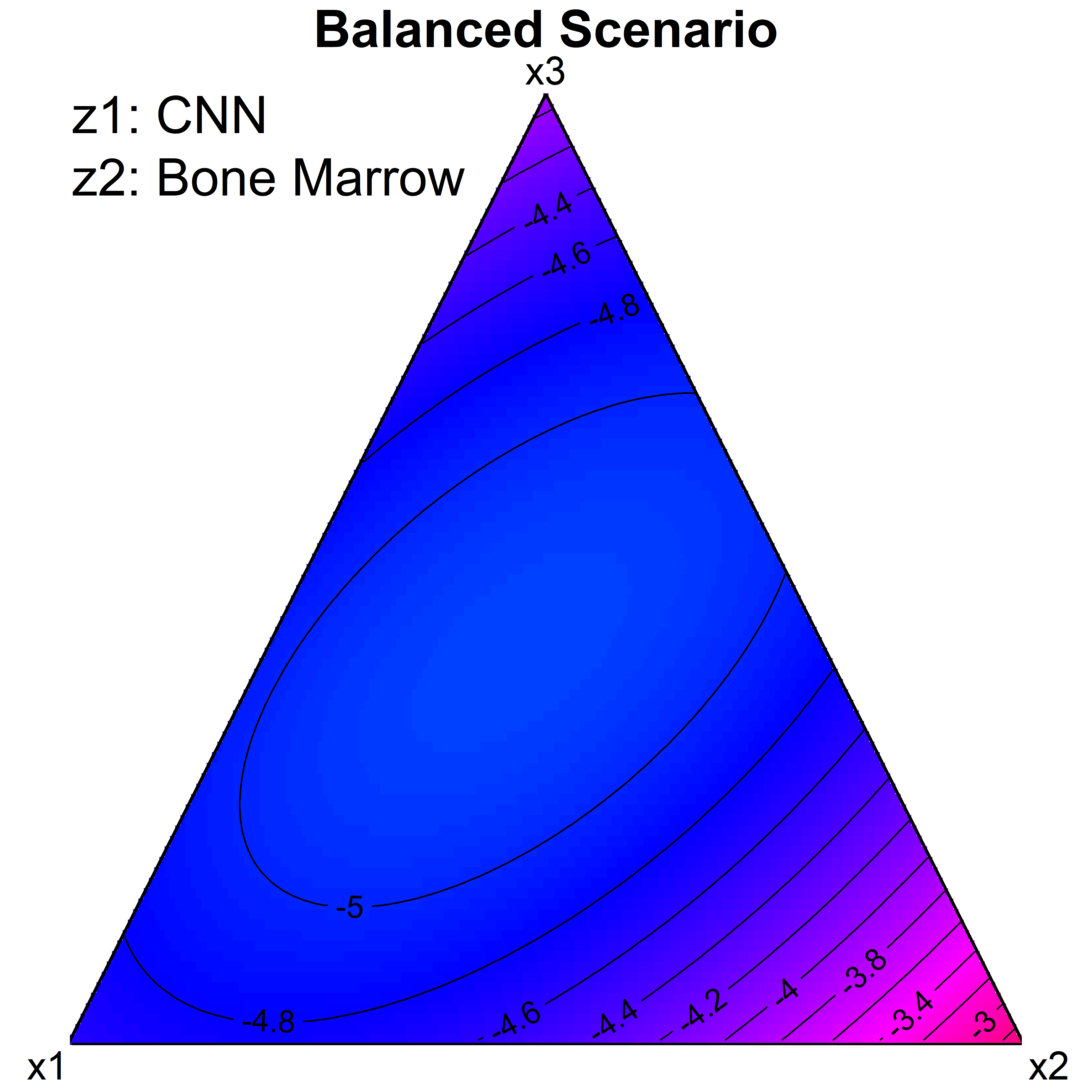}&
\includegraphics[width=.3\textwidth]{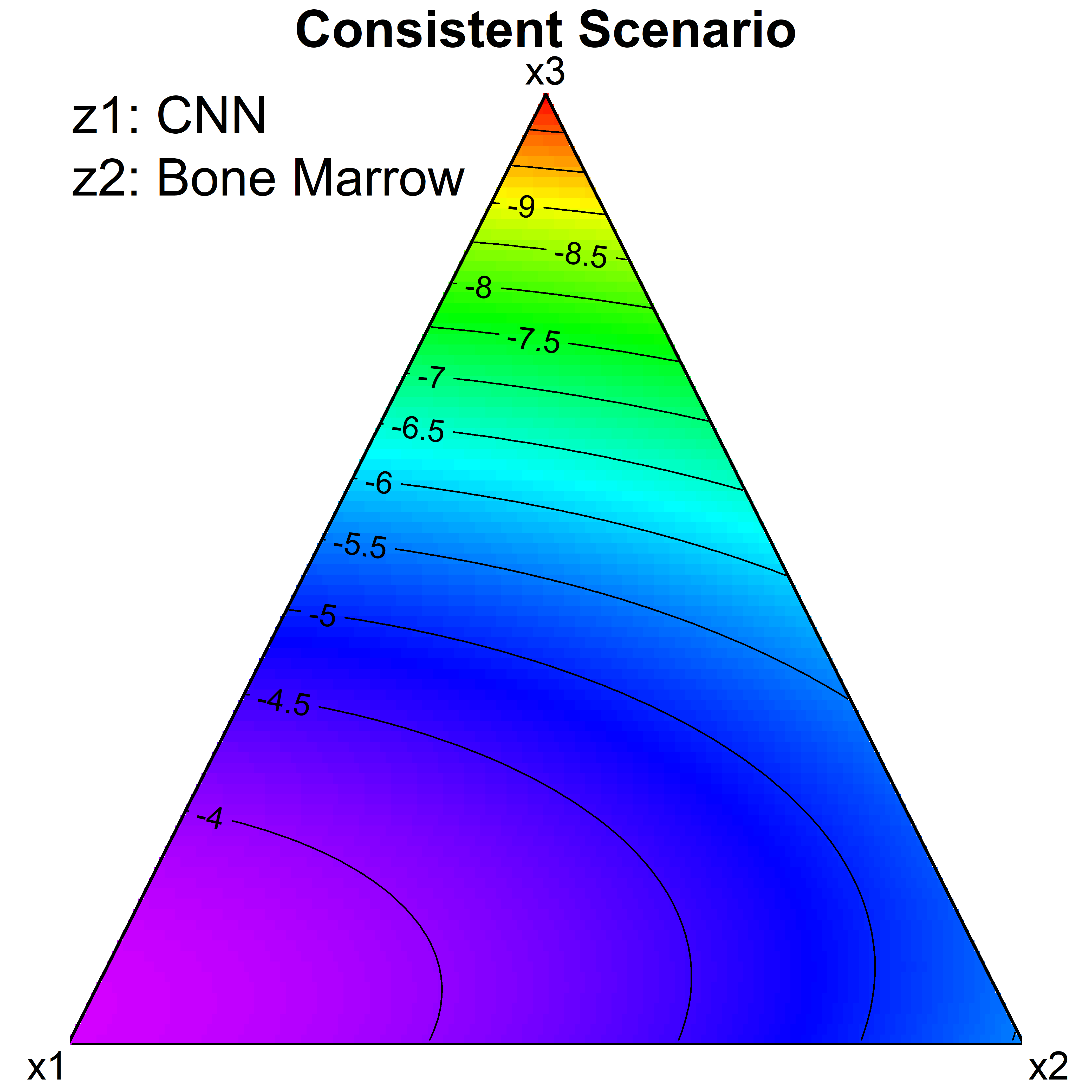}&
\includegraphics[width=.3\textwidth]{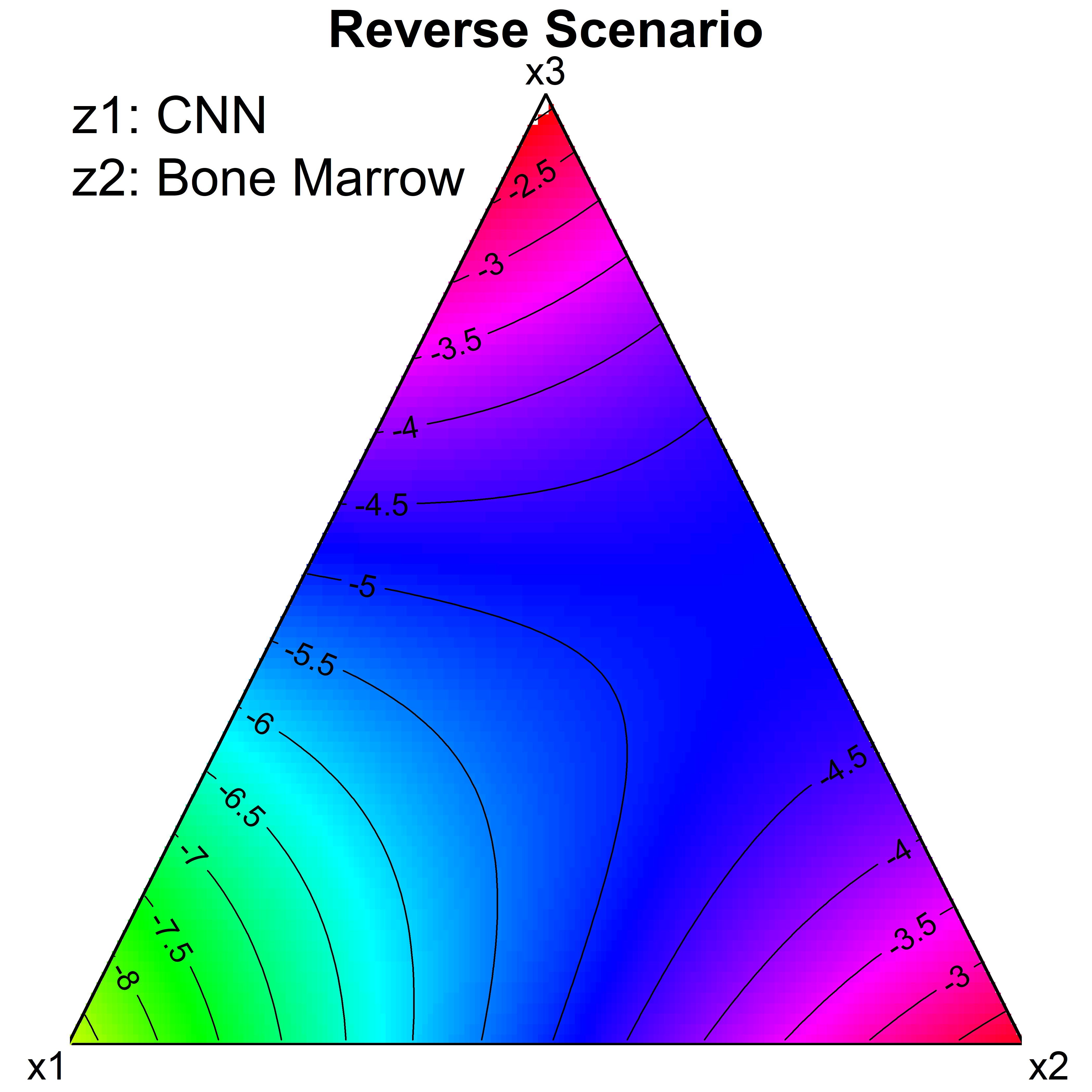}\\
\end{tabular}
\caption{The triangle contour plots of prediction for the Log SD under the three scenarios.}\label{fig:contour-SD}
\end{figure}

Lastly, we examine the impact of predictor variables to the response through the Shapely values in \eqref{eqn:shap-1}-\eqref{eqn:shap-4}.
Table \ref{tab:shap} reports the Shapley values of predictor variables in the estimated models for the mean AUC and the Log SD, respectively.
From the results in the table, it is clear that the proportion of class label 3, $x_{3}$, has the largest impact on the mean AUC for all the three scenarios.
While for the Log SD, the proportion of class label 1, $x_{1}$, has the largest impact to the response under the Reverse and Balanced Scenarios,
and the proportion of class label 3, $x_{3}$, has the largest impact to the response under the Consistent Scenario.
\begin{table}
\begin{center}
\caption{The Shapley values of predictor variables in the estimated model for the mean AUC and the Log SD under the three scenarios.}\label{tab:shap}
\vspace{1.5ex}
\begin{tabular}{lrrr|lrrr}\hline\hline
\multicolumn{4}{c|}{Mean AUC Analysis} & \multicolumn{4}{c}{Log SD Analysis}\\\hline
Coef  & Balanced & Consistent & Reverse & Coef & Balanced & Consistent & Reverse\\\hline
$x_1$       & 0.118 & 0.110 & 0.127 &$x_1$       & \textbf{1.239} & 0.977 & \textbf{2.332} \\
$x_2$       & 0.151 & 0.162 & 0.145 &$x_2$       & 0.753 & 1.533 & 0.605 \\
$x_3$       & \textbf{0.250} & \textbf{0.291} & \textbf{0.213} &$x_3$       & 1.143 & \textbf{3.033} & 0.544 \\
$x_1x_2$    & 0.042 & 0.036 & 0.045 &$x_1x_2$    & 0.226 & 0.106 & 0.112 \\
$x_1x_3$    & 0.046 & 0.043 & 0.049 &$x_1x_3$    & 0.177 & 0.499 & 0.092 \\
$x_2x_3$    & 0.039 & 0.027 & 0.045 &$x_2x_3$    & 0.473 & 0.550 & 0.792 \\
$x_1z_1$    & 0.059 & 0.072 & 0.042 &$x_1z_1$    & 0.301 & 0.024 & 1.156 \\
$x_2z_1$    & 0.039 & 0.037 & 0.041 &$x_2z_1$    & 0.051 & 0.277 & 0.108 \\
$x_3z_1$    & 0.003 & 0.009 & 0.005 &$x_3z_1$    & 0.042 & 0.100 & 0.055 \\
$x_1z_2$    & 0.006 & 0.007 & 0.002 &$x_1z_2$    & 0.403 & 0.449 & 0.264 \\
$x_2z_2$    & 0.017 & 0.025 & 0.004 &$x_2z_2$    & 0.402 & 0.300 & 0.338 \\
$x_3z_2$    & 0.028 & 0.046 & 0.014 &$x_3z_2$    & 0.434 & 0.586 & 0.618 \\
$z_1z_2$    & 0.016 & 0.007 & 0.012 &$z_1z_2$    & 0.229 & 0.751 & 0.261 \\\hline\hline
\end{tabular}
\end{center}
\end{table}

\subsection{Summary of Findings}
Based on the data visualization in Section \ref{sec: data-visualization} and the modeling results in Section \ref{sec: modleing-results},
we provide a summary of the findings as follows.

\begin{inparaitem}
\item The balanced setting of the class label proportions generally gives the best classification accuracy in terms of the mean AUC for all the three test scenarios. This finding is applicable for both the CNN and XGboost algorithms.

\item The proportions of class labels may not be of equal importance for the robustness of the AI algorithms.
For the KEGG and Bone Marrow datasets studied in this paper, the proportion of class label 3 in the training data appears to be more important than the other two class labels for classification accuracy in terms of the mean AUC.

\item The choice of classification algorithms (i.e., covariate variable $z_{1}$) appears to have a statistically significant effect on the response of the mean AUC, but may not be significant for the Log SD, which measures the variation of the AUC values of $m$ classes.

\item The interactions between the class label proportions and the chosen algorithm are often significant for the classification performance in terms of the mean AUC and the Log SD.

\item The predicted mean AUC based on the proposed model generally achieves the largest accuracy under the setting of class proportions being balanced in the training dataset.

\item For the response Log SD, the patterns of significant factors and predicted values are generally more heterogeneous across different level combinations of the covariate variables.
\end{inparaitem}

\section{Concluding Remarks}\label{sec:discussion}
In this work, we propose an experimental design framework to systematically investigate the robustness of the AI classification algorithms.
In particular, we adopt the idea of mixture experiments to investigate how the proportion of class labels in the training dataset affect the classification performance of the AI classification algorithms.
Our investigations have also accommodated the distribution changes on the proportion of class labels between the training dataset and test dataset.
In general, more balance on the class label proportions in the training dataset could give higher classification accuracy for the algorithm.
Although we choose two representative AI classification algorithms, the XGboost and the CNN, with the KEGG and Bone Marrow datasets of three classes in our current study,
the proposed framework can be extended for multiple algorithms and multiple datasets with multiple classes.
For example, this methodology could be extended to investigate the minimal size of a training dataset for various performance objectives given constraints in the balance in the available data.  This research indicates that instead of random splits of training, validation, and test, 
one might achieve a more robust algorithm by selecting the training set to be balanced across the classes.

This research also shows the value of building a statistical regression model to determine not only the significant factors affecting algorithm performance, but also to provide an analytical tool for predicting the mean AUC and Log SD values for a future test sets with previously unseen proportions of class labels.
Moreover, our proposed framework is also suitable to investigate other characteristics of AI algorithms, such as the AI fairness and the AI reliability \citep{freeman2019challenges, hong2018big}.


There are a few directions for the future work.
First, it will be interesting to study an optimal design of mixture experiments for the investigation of AI classification algorithms, especially when the number of class is large.
Note that the classical optimal design, such as the I-optimal design, is often based on certain statistical models \citep{goos2016optimal, li2018ei}.
For the investigation of the AI classification algorithm, the design optimality will be associated with both the statistical model and AI algorithms.
Second, when the dataset used in the AI classification algorithms involve many classes, it is also challenging to design a good mixture experiment with a small run size.
One possible solution is to consider the batch sequential experiments and active learning \citep{deng2009active} with the focus on the proportions of class labels dominated by a small number of class labels in each stage.
Third, our current analysis of the experiment outcomes is based on linear models. While the experiment outcomes can also be categorical.
It will be interesting to adopt other flexible modeling methods for the analysis such as the Gaussian process modeling \citep{deng2017additive}.



 \section*{Acknowledgement}
The authors acknowledge the Commonwealth Cyber Initiative (CCI) for providing computational resources. This research was a part of Security and Software Engineering Research Center (S2ERC) and was supported by National Science Foundation under Grant CNS-1650540 to Virginia Tech. The work by Hong was also partially supported by the Virginia Tech Institute for Critical Technology and Applied Science (ICTAS) Junior Faculty Award.

\appendix
\section{The Shapley Formula Under the Linear Model}
Suppose that one considers a linear regression model with $p$ predictor variables $x_{1}, \ldots, x_{p}$.
Denote the $S_{\textrm{all}} = \{1, \ldots, p\}$ to a full index set of predictor variables in the model.
That is, the regression model can be written as
\begin{align*}
y|\bm x = f(x_{1}, \ldots, x_{p}) + \epsilon = \sum_{j \in S_{\textrm{all}}}  \beta_{j} x_{j} + \epsilon.
\end{align*}
Denote the $M \subseteq \{1, \ldots, p\}$ to be an index subset of variables.
Let $S_{-j} = \{1, \ldots, j-1, j+1, \ldots, p\}$ = $S_{\textrm{all}}  \setminus \{j\}$, which is the full index set except the index $j$.
The Shapley value is to examine the impact of the predictor variable $X_{k}$ based on the idea of the cooperative game theory \citep{shapley1953value}.
According to the Shapley value, the impact of variable $x_{k}$ is
\begin{align}
\phi_{k} = \sum_{M \subseteq S_{-k} } w_{M} \left[ \nu(M \cup \{k\}) - \nu(M) \right],
\end{align}
where $\nu(\cdot)$ is a metric of information gain or utility function. The $w_{M}$ a weight is as $1/(p \binom{p-1}{q})$ with $q = \textrm{card}(M)$.
In the context of regression, one popular choice is $\nu(M) = \Exp(y|\bm x_{M})$,
which is the expected output of the predictive model, conditional on the predictor values $\bm x_{M} = \{x_{j}: j \in M \}$ of this subset.
Note that the expectation here is with respect to both $y$ and $\bm x_{\bar{M}}$ where $\bm x_{\bar{M}} = \{ x_{j}: j \notin M \}$.
Specifically, we have
\begin{align*}
    \nu(M)  = \Exp(y|\bm x_{M}) & = \Exp_{\bm x_{\bar{M}}}  \left [ \Exp_{y | \bm x} \left( \sum_{j \in M} \beta_{j} x_{j}  +  \sum_{j \in \bar{M} } \beta_{j} x_{j} + \epsilon \right)  \right] \nonumber \\
    & = \Exp_{\bm x_{\bar{M}}} \left ( \sum_{j \in M} \beta_{j} x_{j}  +  \sum_{j \in \bar{M} } \beta_{j} x_{j}   \right)
     = \sum_{j \in M} \beta_{j} x_{j}  +  \sum_{j \in \bar{M} } \beta_{j} \Exp(x_{j}).
\end{align*}
Consequently, we have
\begin{align*}
    \nu(M \cup \{k\}) - \nu(M) &= \sum_{j \in M \cup \{ k\} } \beta_{j} x_{j}  +  \sum_{j \in \bar{M} \setminus \{k\} } \beta_{j} \Exp(x_{j})
    - \left[ \sum_{j \in M} \beta_{j} x_{j}  +  \sum_{j \in \bar{M} } \beta_{j} \Exp(x_{j}) \right] \nonumber \\
    & = \beta_{k} x_{k} - \beta_{k} \Exp(x_{k}).
\end{align*}
Noting that the sum of the weights is 1, we thus have $\phi_{k} =   \beta_{k} [ x_{k} - \Exp(x_{k}) ]$.

\bibliographystyle{apalike}
\bibliography{Ref_Deep}
\end{document}